\documentclass{article}

\usepackage[square,numbers,compress]{natbib}
\usepackage[preprint]{arxiv}

\usepackage[utf8]{inputenc} 
\usepackage[T1]{fontenc}    
\usepackage{hyperref}       
\usepackage{url}            
\usepackage{booktabs}       
\usepackage{amsfonts}       
\usepackage{nicefrac}       
\usepackage{microtype}      
\usepackage{xcolor}         

\usepackage{amssymb}
\usepackage{multirow}
\usepackage{arydshln}
\usepackage{wrapfig}
\usepackage{amsmath}
\usepackage{colortbl}
\usepackage{enumitem}
\usepackage{adjustbox}

\usepackage{tcolorbox}

\usepackage{makecell}
\usepackage{color}
\usepackage{pgfplots}
\usepackage{circledtext}
\usetikzlibrary{shapes}
\usetikzlibrary{arrows,decorations.pathmorphing,backgrounds,positioning,fit,petri}
\usetikzlibrary{arrows.meta,fit,shapes.arrows}
\usetikzlibrary{positioning,shadows,patterns}
\usepackage{caption}

\usepackage{amssymb}

\definecolor{tiffanyblue}{RGB}{129,216,208}
\definecolor{bangdiblue}{RGB}{0,149,182}
\definecolor{kleinblue}{RGB}{0,47,167}

\usepackage{tikz}
\usepackage{subfig}
\usepackage{tkz-kiviat,pgfplots}
\pgfplotsset{compat=newest}
\usepgfplotslibrary{fillbetween}
\usepgfplotslibrary{patchplots} 

\usepackage{listings}

\usepackage{marvosym}

\newtcolorbox{promptbox}[2][]{
	width=\linewidth,
	colback = gray!8, 
	colframe = gray!8, 
	boxsep=0pt,left=5pt,right=5pt,top=2pt,bottom=2pt,
	fontupper=\linespread{1.2}\selectfont,
	title=#2,#1,
        fontupper=\small}

\definecolor{given}{RGB}{197,217,197}
\definecolor{response}{RGB}{176,224,230}
\definecolor{myBrown}{RGB}{140, 50, 50}
\definecolor{myPurple}{RGB}{128, 0, 200} 

\usepackage{pifont}
\newcommand{\greentick}{\textcolor{green!70!black}{\ding{52}}}
\newcommand{\redcross}{\textcolor{red}{\ding{56}}}

\definecolor{forestgreen}{RGB}{34, 139, 34}
\definecolor{firebrick}{RGB}{178, 34, 34}

\usepackage{bm}

\newcommand{\uparr}{$\color{forestgreen}{\bm{\uparrow}}$}
\newcommand{\downarr}{$\color{firebrick}{\bm{\downarrow}}$}

\newcommand\name{\texttt{CultureForest}}

\def\huggingface{\raisebox{-1.5pt}{\includegraphics[height=1.05em]{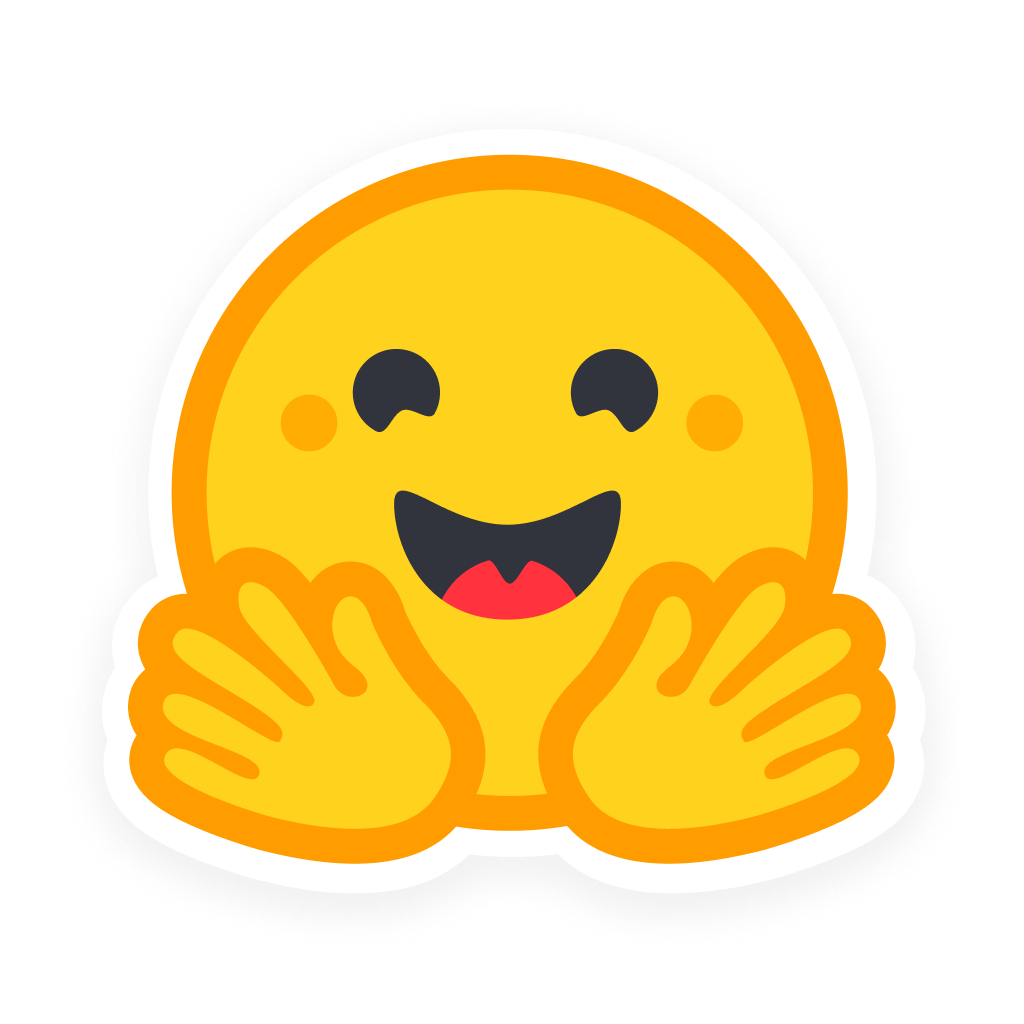}}}
\def\github{\raisebox{-1.5pt}{\includegraphics[height=1.05em]{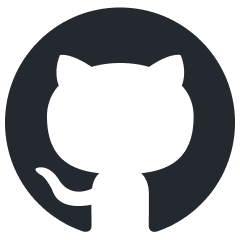}}}

\newcommand{\ghlink}{https://github.com/YYF-Tommy/CultureForest}
\newcommand{\hflink}{https://huggingface.co/datasets/TommyYe/CultureForest}

\title{CultureForest: Understanding and Evaluating Cultural Norm Grounded Reasoning in LLMs}

\author{%
  Yangfan Ye$^{1}$\quad
  Xiaocheng Feng$^{1}$$^{\textrm{\Letter}}$\quad
  Jialong Tang$^{\textrm{\Letter}}$\quad
  Xiayu Cao$^{1}$\quad
  Zihan Zhang$^{3}$\\
  \textbf{Xiachong Feng$^{2}$\quad
  Baosong Yang\quad
  Bing Qin$^{1}$}\\
  $^{1}$Harbin Institute of Technology\quad
  $^{2}$The University of Hong Kong\quad
  $^{3}$Harvard University\\
  \texttt{\{yfye,xcfeng\}@ir.hit.edu.cn}
  \\
  \github~~\url{\ghlink}\\
  \huggingface~~\url{\hflink}\\
}

\begin{document}

\maketitle

\vspace{-\baselineskip}

\begin{abstract}
Existing research largely reduces cultural intelligence in LLMs to a knowledge-level problem, overlooking whether models can effectively utilize their acquired knowledge in realistic scenarios.
To bridge this gap, we introduce {\name}, a benchmark for \textit{Cultural Norm Grounded Reasoning}. Each question is grounded in a small set of atomic norms, enabling verifiable and attributable evaluation. {\name} comprises 5,378 examples across 8 domains and 53 countries/regions, and supports a progressive evaluation from multiple-choice to open-ended generation.
Extensive experiments reveal that even top-tier models degrade substantially in open-ended settings, accompanied by pronounced cross-region disparities. Through targeted analysis, we uncover several consistent patterns: (1) test-time reasoning yields limited gains and may exacerbate inequity; (2) models exhibit highly shared regional preference structures; (3) model responses are markedly conservative, especially under stricter cultural constraints; and (4) by disentangling cultural knowledge acquisition from cultural reasoning, we show that while LLMs possess substantial cultural knowledge, their performance is further bottlenecked by its effective use.
These findings point to a necessary shift from knowledge-centric evaluation toward measuring knowledge-grounded reasoning.
\end{abstract}

\section{Introduction}\label{sec:intro}

From cross-cultural communication to region-specific decision-making, cultural intelligence is increasingly critical for deploying large language models (LLMs) in real-world, globally diverse settings~\citep{li2024culturellm, li2024culturepark, naous-etal-2024-beer, ye-etal-2024-globesumm, ye2026x1, yuan2026culture}. 
In such scenarios, models are expected to interact with users from diverse cultural backgrounds, where seemingly minor behavioral choices, such as forms of politeness, social etiquette, or taboo avoidance, can carry significant implications~\citep{durham1976adaptive, bulte2025llms}.

To assess such capabilities, a growing body of work has introduced culture-related benchmarks. Prior efforts such as CulturalBench~\citep{chiu-etal-2025-culturalbench}, FORK~\citep{palta-rudinger-2023-fork} and TyDiP~\citep{srinivasan-choi-2022-tydip} primarily evaluate models' understanding of cultural knowledge through multiple-choice or classification tasks, focusing on domains such as everyday life, food, and politeness. More recent benchmarks, including NativQA~\citep{hasan-etal-2025-nativqa} and BLEnD~\citep{myung2024blend}, extend evaluation to short-form question answering settings, while MultiNRC~\citep{fabbri2025multinrc} introduces reasoning tasks embedded in cultural contexts (e.g., traditional reasoning or math reasoning framed within cultural narratives). A comparison of representative benchmarks is summarized in Table~\ref{tab:comparision}.
Despite these advances, existing evaluations share two fundamental limitations:

\textbf{(1)} Existing evaluations largely reduce cultural intelligence to a \emph{knowledge level} problem. Models are primarily tested on whether they can recall related cultural norms. While important, this paradigm overlooks a key aspect: \emph{can models effectively utilize acquired cultural knowledge for reasoning?} In practice, failures often arise not from missing knowledge, but from misapplying it.

\textbf{(2)} Most benchmarks predominantly rely on \emph{closed-form evaluation} (e.g., multiple-choice or classification), which diverges from real-world open-ended scenarios and often exposes answer candidates in context, thereby simplifying the task and potentially inflating performance estimates.

To bridge these gaps, we introduce \textbf{{\name}}, a benchmark for \emph{Cultural Norm Grounded Reasoning}, designed to evaluate whether models can move beyond knowledge recall toward effective knowledge utilization. {\name} is built upon the following key principles:
\begin{itemize}[leftmargin=*]
\setlength{\parsep}{0pt}
\setlength{\parskip}{0pt}
\item \textbf{Broad Domain and Country/Region Coverage.}  
{\name} spans 8 domains of everyday cultural practices and covers a diverse set of 53 countries/regions across major geographic areas. This broad coverage enables systematic evaluation of models' performance across heterogeneous cultural contexts, supporting fine-grained analysis of cross-region disparities and robustness.
\item \textbf{Norm-Grounded and Attributable Design.}  
Each question is explicitly grounded in a small set of atomic cultural norms, enabling \textbf{verifiable} evaluation and \textbf{attributable} analysis. This structure allows model predictions to be traced back to specific norms, making it possible to disentangle knowledge acquisition and reasoning.
\item \textbf{Progressive Evaluation from Closed- to Open-Form.}  
{\name} provides three difficulty levels (Easy, Medium, Hard) that progressively reduce external guidance, ranging from multiple-choice selection to open-ended generation. This setup enables a systematic evaluation of model behavior under increasingly realistic and unconstrained scenarios.
\item \textbf{Scalable and Robust Open-Ended Evaluation.}  
To support evaluation in open-ended settings, we develop a lightweight verifier, achieving strong effectiveness and generalization. This enables consistent and scalable assessment of norm compliance beyond closed-form settings.
\end{itemize}

We conduct extensive experiments across a diverse set of models and scales, revealing fundamental limitations of current LLMs in open-ended cultural scenarios, along with pronounced cross-region disparities.
Through multi-faceted analysis, including \textit{scaling effects (model size and test-time reasoning)}, \textit{regional preference structures}, and \textit{distributional patterns of cultural norm compliance}, we uncover systematic weaknesses that remain largely hidden under conventional evaluation settings.
By disentangling cultural knowledge acquisition from cultural reasoning, we find that although modern LLMs possess substantial cultural knowledge, their performance is bottlenecked by its effective use.
We therefore argue that advancing cultural intelligence requires moving beyond knowledge-centric evaluation. {\name} takes a step in this direction, offering a diagnostic testbed for studying knowledge-grounded reasoning in cultural settings. (\text{see Figure~\ref{fig:example} for an \textbf{example} of {\name}.})

\begin{table*}[t]
    \centering
    \scriptsize
    \caption{Comparison between {\name} and existing other culture-related benchmarks. Most prior benchmarks primarily evaluate models' acquisition of cultural (often factual) knowledge. Detailed domain coverage information for each benchmark is provided in Appendix~\ref{app:domain} Table~\ref{tab:domain}.}
    \begin{adjustbox}{width=\textwidth}
        \begin{tabular}{cccccc}
            \toprule
            \textbf{Benchmarks} & \textbf{Domain} & \textbf{Evaluation Focus} & \textbf{Open-Ended}  & \textbf{\# Country/Region} & \textbf{\# Size} \\
            \midrule
            CulturalBench~\citep{chiu-etal-2025-culturalbench} & \makecell{Multi-domain (Everyday-life)} & Cultural Norms / Knowledge & {\redcross} & 45 & 1,696 \\
            FORK~\citep{palta-rudinger-2023-fork} & Food & Cultural Norms / Knowledge & {\redcross} & 10 (US vs. Non-US) & 184 \\
            TyDiP~\citep{srinivasan-choi-2022-tydip} & Politeness & Politeness Discrimination & {\redcross} & >=9 & 4,500 \\
            NativQA~\citep{hasan-etal-2025-nativqa} & \makecell{Multi-domain (Everyday-life)} & Cultural Norms / Knowledge & {\greentick} (Short-form) & 9 & $\sim$64,000 \\
            BLEnD~\citep{myung2024blend} & \makecell{Multi-domain (Everyday-life)} & Cultural Norms / Knowledge & {\greentick} (Partial \& Short-form) & 16 & $\sim$52,600 \\
            MultiNRC~\citep{fabbri2025multinrc} &  \makecell{Linguistic, Wordplay, Tradition, Math} & Reasoning in Cultural Context & {\greentick} (Short-form) & >=3 & 1,055 \\
            \midrule
            {\name} (our) & \makecell{Multi-domain (Everyday-life)} & \makecell{Cultural Norm Grounded Reasoning} & {\greentick} (Partial \& \textbf{Long-form}) & 53 & 5,378 \\
            \bottomrule
        \end{tabular}
    \end{adjustbox}
    
    \label{tab:comparision}
\end{table*}

\section{CultureForest Benchmark}\label{sec:benchmark}

We formulate \textbf{Cultural Norm Grounded Reasoning} as the core task of {\name}: given a contextualized scenario, models must reason about and apply acquired cultural knowledge to determine or generate appropriate behavior. 
To enable principled evaluation, each question is explicitly grounded in a small set of \textbf{atomic cultural norms}, ensuring that model behaviors are both \textbf{Verifiable} (objectively assessable) and \textbf{Attributable} (traceable to specific cultural norms).

\subsection{Data Source and Norm Curation}\label{sec:norm_curation}
We source cultural norms from Cultural Atlas~\citep{culturalatlas}, a platform providing expert-verified descriptions of societal attitudes and behaviors across diverse regions, grounded in both community-sourced insights and academic review. 
We reorganize the data into a structured format:
\begin{center}
\fbox{%
\scriptsize
\parbox{0.95\linewidth}{%
\texttt{Norm Bundle = \{``country'': ``Afghanistan'', ``domain'': ``Business Culture'', ``topic'': ``Meetings'', ``norm'': ``Bargaining is an acceptable and common way of negotiating.''\}}
}%
}
\end{center}
Each question $q_i$ in {\name} is grounded in a set of atomic cultural norms $S_{q_i} = \{n_1, \dots, n_L\}$, serving as its underlying knowledge basis. To ensure quality, diversity, and applicability, the norms within each $S_{q_i}$ are selected according to the following principles:

\begin{itemize}[leftmargin=*]
\setlength{\parsep}{0pt}
\setlength{\parskip}{0pt}
\item \textbf{Scenario Consistency.}
Norms are restricted to a shared pool $\operatorname{Pool}(c,d,t)$ defined by identical \texttt{country} ($c$), \texttt{domain} ($d$), and \texttt{topic} ($t$), ensuring joint applicability within a coherent scenario.

\item \textbf{Content Diversity.}
To avoid semantic redundancy, each norm $n \in \operatorname{Pool}(c,d,t)$ is embedded\footnote{Using \texttt{text-embedding-3-large} as the embedding model.} via:
\begin{center}
\fbox{%
\scriptsize
\parbox{0.95\linewidth}{%
\texttt{``Behavioral constraint imposed by the following cultural norm: \{norm\}''}
}%
}
\end{center}
We then perform Gaussian Mixture Model (GMM) clustering and require selected norms to (i) belong to distinct clusters and (ii) achieve an average silhouette score above a threshold $\tau$ (defined as the pool-level mean), ensuring semantic heterogeneity.
\item \textbf{Norm Strictness.}
Cultural norms differ in their degree of strictness\footnote{For example, in Muslim communities, consuming pork is strictly prohibited; whereas in China, though using chopsticks is customary, using forks in certain contexts (e.g., steakhouse) is sometimes acceptable.}. 
To quantify this property, we estimate the acceptability of violating each norm via the following classification task:

\begin{center}
\fbox{%
\scriptsize
\parbox{0.95\linewidth}{%
\texttt{``In \{country\} culture, is it ever acceptable to violate the following cultural norm? \{norm\} (Yes/Sometimes/No)''}
}%
}
\end{center}
Aggregated responses yield \textit{Yes}, \textit{Sometimes}, and \textit{No}. We discard norms universally labeled \textit{Yes}, and require each $S_{q_i}$ to include at least one strict (\textit{No}) norm, ensuring non-trivial constraints and avoiding degenerate cases. (Implementation details are provided in Appendix~\ref{app:strictness})
\end{itemize}

\subsection{Benchmark Construction}

\begin{figure*}[ht]
    \centering
    \includegraphics[width=\textwidth]{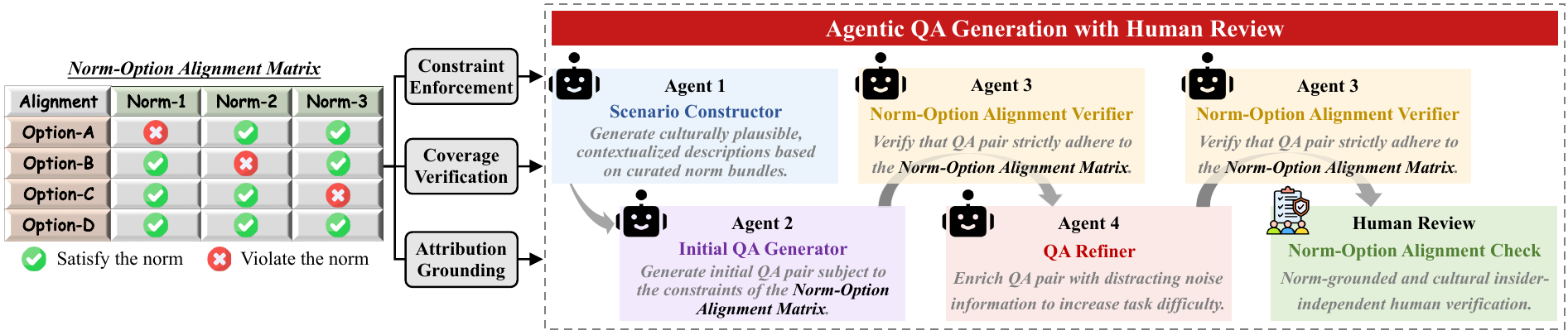}
    \caption{Overview of the norm-option alignment constraints and agentic QA Generation pipeline.}
    \label{fig:QA_gen}
\end{figure*}

We construct {\name} via an agentic QA generation pipeline with human verification. Each question $q_i$ is grounded in a set of atomic cultural norms $S_{q_i}$, where the set size $|S_{q_i}|$ is fixed to $\mathbf 3$ to intentionally align with the \emph{norm-option alignment matrix} illustrated in Figure~\ref{fig:QA_gen}.

Following the pipeline in Figure~\ref{fig:QA_gen}, answer options are designed to reflect different patterns of norm satisfaction. Specifically, we enforce that each option corresponds to a distinct combination of satisfied and violated norms, ensuring that correct reasoning requires integrating all norms rather than relying on partial cues. 
To guarantee the faithfulness between options and their intended norm constraints, we employ an automatic \emph{alignment checker} that verifies whether each option satisfies or violates the designated subset of norms.
The generation process further incorporates refinement and human validation to ensure both correctness and appropriate difficulty.
The generation pipeline is implemented using \textit{GPT-4.1-2025-04-14}.
Detailed prompt designs and annotation procedures are provided in Appendix~\ref{app:benchmark}. The high inter-annotator agreement and exceptional compliance rates observed during manual inspection underscore the superior quality of \name.

\subsection{Difficulty Level Partition and Evaluation}

We design 3 difficulty levels in {\name} to progressively reduce explicit external guidance:.

\paragraph{(1) Easy.}
A multiple-choice setting where models select the correct answer from candidate options. Before shuffling, option D is the correct answer, as it satisfies all cultural norms as shown in Figure~\ref{fig:QA_gen}.

\paragraph{(2) Medium.}
Models are required to independently assess the cultural appropriateness of each option via binary judgments (\texttt{True} / \texttt{False}). According to the norm-option alignment matrix in Figure~\ref{fig:QA_gen}, options A,B,C correspond to culturally inappropriate behaviors, while option D is appropriate. A prediction is considered correct only if all options are classified correctly.

For both \textbf{Easy} and \textbf{Medium} settings, we report Accuracy as the evaluation metric.
\paragraph{(3) Hard.}
An open-ended generation setting where no options are provided. Models must produce responses that appropriately comply with the underlying cultural norms.

\paragraph{Verifier Model for Evaluation.}
We train a lightweight verifier model (\texttt{C-Verifier}) based on \textit{Qwen3.5-0.8B-Base}, enabling efficient and scalable evaluation in open-ended settings. Given an input pair \textit{(response, norm)}, \texttt{C-Verifier} classifies the response into \textit{Satisfy}, \textit{Neutral}, or \textit{Violate} based on its compliance with the specified norm.
The training data consists of two complementary sources:
\begin{itemize}[leftmargin=*]
\setlength{\parsep}{0pt}
\setlength{\parskip}{0pt}
\item $\text{Data}_{\text{syn}}$: \textbf{synthetic supervision} derived from benchmark construction, where each option is paired with its associated norms and labeled as \textit{Satisfy} or \textit{Violate} (Figure~\ref{fig:QA_gen}). \textit{Neutral} labels are constructed from unrelated norm-option pairs. The data is split into train and valid sets with a 9:1 ratio.
\item $\text{Data}_{\text{real}}$: \textbf{human-annotated responses} sampled from diverse models, covering a wide range of model families and capability levels (specifically, the seven models listed under the ``Real Scenarios'' section of Table~\ref{tab:k-fold}), thereby improving robustness to real-world scenarios. The human annotation protocol follows the procedure detailed in Appendix~\ref{app:benchmark}, with the addition of a \textit{Neutral} label to capture responses that are neither strictly compliant nor explicitly violating.
\end{itemize}

\paragraph{Validation.}
To assess the effectiveness and generalization of \texttt{C-Verifier}, we conduct validation on $\text{Data}_{\text{syn}}$ (valid) and adopt a K-fold cross validation (leave-one-model-out) on the subsets of $\text{Data}_{\text{real}}$ to evaluate generalization. In each fold, we exclude all responses generated by one specific target model from the training set $\text{Train}_{\text{real}}$ and use them exclusively for evaluation. This ensures that the verifier is tested on unseen model behaviors, thereby rigorously evaluating its cross-model generalization capability.
As shown in Appendix~\ref{app:k-fold} and Table~\ref{tab:k-fold}, \texttt{C-Verifier} consistently outperforms strong LLM-as-a-judge baselines, demonstrating superior robustness and cross-model generalization.

\paragraph{Scoring.}
For each generated response $x_q$ to question $q$ and a given cultural norm $n \in S_q$, 
the \texttt{C-Verifier} produces logits over three labels 
$\mathcal{Y} = \{\textit{Satisfy}, \textit{Neutral}, \textit{Violate}\}$. 
We convert these logits into a probability distribution via softmax, denoted as 
$p(y \mid x_q, n)$ for $y \in \mathcal{Y}$.

We then define a utility function $w: \mathcal{Y} \rightarrow \mathbb{R}$ to quantify norm compliance. 
\textit{Satisfy} and \textit{Violate} are assigned reward $+1$ and penalty $-1$, respectively, while 
\textit{Neutral} responses reflect uncertainty or avoidance and are therefore assigned a moderate penalty $-\lambda$, where $\lambda \in (0,1)$:
{
\footnotesize
\begin{equation}
w_{\text{No}}(y) =
\begin{cases}
+1, & y = \textit{Satisfy}, \\
-\lambda, & y = \textit{Neutral}, \\
-1, & y = \textit{Violate}.
\end{cases}
\quad w_{\text{Sometimes}}(y) = \alpha \cdot w_{\text{No}}(y),
\end{equation}
}
To account for different levels of norm strictness (\S\ref{sec:norm_curation}), we scale the utility for \textit{Sometimes} (sometimes acceptable to violate) norms, where $\alpha \in (0,1)$ controls the relative strength.

Given a question $q \in \mathcal{Q}$ associated with a set of norms $S_q = \{n_1, n_2, n_3\}$, 
we compute the expected utility for each norm and average across norms, where $w_n$ is determined by the norm type (\textit{No} or \textit{Sometimes}). And the score is obtained by averaging over all questions:
{
\footnotesize
\begin{equation}
\operatorname{Score}(q) = \frac{1}{|S_q|} \sum_{n \in S_q} \sum_{y \in \mathcal{Y}} w_n(y) \cdot p(y \mid x_q, n),
\quad \operatorname{Score}_{\text{open}} = \frac{1}{|\mathcal{Q}|} \sum_{q \in \mathcal{Q}} \operatorname{Score}(q).
\end{equation}
}
To ensure comparability, we linearly normalize scores into $[-100,100]$ using theoretical bounds:
{
\footnotesize
\begin{equation}
\widetilde{\operatorname{Score}} = -100 + 200 \cdot \frac{\operatorname{Score}_{\text{open}} - L}{U - L},
\end{equation}
}
where $L$ and $U$ denote the minimum and maximum achievable scores, respectively.

\paragraph{Sensitivity Analysis.}
We perform a grid search over $\lambda, \alpha \in \{0.1, 0.3, 0.5, 0.7, 0.9\}$ and evaluate robustness via pairwise rank correlations under different parameter choices, 
including both model-level and country-level rankings. 
As shown in Appendix~\ref{app:sensitivity} and Figure~\ref{fig:sensitivity}, consistently high correlations indicate that our conclusions and rankings are stable across parameter configurations.
We therefore adopt $\lambda = \alpha = 0.5$, as this configuration achieves the highest overall correlation.

\subsection{Benchmark Statistics}

{\name} comprises 5,378 instances spanning 8 domains and 53 countries/regions, covering diverse geographic areas including Asia, Europe, the Middle East, Africa, Oceania, and the Americas. 
Each instance is paired with three parallel evaluation formats: \textbf{Easy} (multiple-choice), \textbf{Medium} (independent binary classification), and \textbf{Hard} (open-ended generation), enabling a controlled assessment of model performance under progressively reduced external guidance.
Detailed statistics on domain and geographic distributions are provided in the Appendix~\ref{app:stat} and Figure~\ref{fig:stat}.

\section{Experiments}\label{sec:analysis}

\subsection{Setup}

\paragraph{Models.}
We evaluate a variety of both top-tier reasoning/non-reasoning models and smaller-scale models.
The complete model list, snapshot versions, and citations are provided in Appendix~\ref{app:models}.
\paragraph{Evaluation Protocol.}
We set the maximum generation length to 65,536 tokens and use sampling with $T=0.7$ and $p=0.95$. 
For models with fixed or non-configurable settings, we adopt their default configurations.
For performance, we report \textbf{Mean@3} and \textbf{Pass@3}. 
To quantify cross-region disparity, we report the standard deviation (\textbf{Std}) and the max--min gap (\textbf{Gap}) over region-level scores. The main results on {\name} across three difficulty levels are reported in Table~\ref{tab:main}.

\begin{table*}[t]
    \centering
    \scriptsize
    \caption{Main results on {\name} across three levels. $\bigstar$ indicates the model used in the data construction pipeline. \textbf{Bold} denotes the best performance, and \underline{underline} denotes the second-best.}

    \begin{adjustbox}{width=\textwidth}
        \begin{tabular}{lcccc|cccc|cccc}
            \toprule
            \multirow{4}{*}{Models} & \multicolumn{4}{c}{Easy} & \multicolumn{4}{c}{Medium} & \multicolumn{4}{c}{Hard} \\
            \cmidrule(lr){2-5}
            \cmidrule(lr){6-9}
            \cmidrule(lr){10-13}
            & \multicolumn{2}{c}{\makecell{Performance {\uparr}}} & \multicolumn{2}{c}{\makecell{Disparity {\downarr}}}  & \multicolumn{2}{c}{\makecell{Performance {\uparr}}} & \multicolumn{2}{c}{\makecell{Disparity {\downarr}}}  & \multicolumn{2}{c}{\makecell{Performance {\uparr}}} & \multicolumn{2}{c}{\makecell{Disparity {\downarr}}} \\
            \cmidrule(lr){2-3}
            \cmidrule(lr){4-5}
            \cmidrule(lr){6-7}
            \cmidrule(lr){8-9}
            \cmidrule(lr){10-11}
            \cmidrule(lr){12-13}
            & Mean@3 & Pass@3 & Std & Gap & Mean@3 & Pass@3 & Std & Gap & Mean@3 & Pass@3 & Std & Gap \\
            \midrule
            \multicolumn{13}{c}{\textbf{Top-Tier Non-Reasoning Models}} \\
            \midrule
            $\bigstar$ GPT-4.1-2025-04-14 & 94.38 & 95.50 & 4.12 & 20.87 & 40.15 & 47.62 & 10.33 & 50.44 & 52.15 & 62.08 & 16.87 & 80.85 \\
            GPT-5.2-2025-12-11 & 92.35 & 93.77 & 3.67 & 14.83 & \textbf{53.25} & 60.21 & 9.31 & 46.45 & 51.29 & 62.44 & 17.28 & 82.61 \\
            Deepseek-V3 & 93.97 & 94.79 & 3.54 & 18.02 & 25.02 & 27.95 & 7.71 & 29.48  & 50.73 & 60.41 & 17.19 & 82.32 \\
            Llama-3.3-70B-Instruct & \underline{95.97} & 96.28 & 2.75 & 11.27 & 31.42 & 33.23 & 9.97 & 39.82 & 34.33 & 45.99 & 20.76 & 89.46 \\
            Llama-4-Scout-17B-16E-Instruct & 92.74 & 94.18 & 3.68 & 15.24  &  44.37 & 46.84 & 8.77 & 39.91 & 33.38 & 45.49 & 19.39 & 87.45 \\
            Claude-Sonnet-4 & 95.23 & 95.63 & 3.08 & 15.90 & \underline{52.83} & 54.70 & 10.56 & 48.24 & 48.03 & 58.33 & 16.80 & 79.22 \\
            Qwen2.5-72B-Instruct & 92.96 & 94.29 & 3.83 & 15.00 & 27.71 & 31.37 & 8.54 & 34.59 & 42.26 & 54.38 & 17.70 & 87.48 \\
            
            \midrule
            \multicolumn{13}{c}{\textbf{Top-Tier Reasoning Models}} \\
            \midrule
            OpenAI-o3 & 91.51 & 94.31 & 4.49 & 17.53 & 50.61 & \underline{62.33} & 9.89 & 42.71 &  \textbf{59.89} & \textbf{71.02} & 14.70 & 72.31 \\
            Deepseek-V3.2-Thinking & 93.65 & 96.04 & 4.32 & 18.97 & 42.44 & 53.61 & 8.98 & 37.06 &  47.91 & 59.03 & 17.62 & 82.49 \\
            Gemini-3.1-Pro-Preview & 91.58 & 94.66 & 6.15 & 29.21 & 52.22 & 62.03 & 11.36 & 55.36 & 56.37 & 66.11 & 16.14 & 75.66 \\
            Claude-Sonnet-4-Thinking & 94.29 & \underline{96.91} & 3.35 & 14.36 & 49.28 & \textbf{62.38} & 9.12 & 37.58 & 45.48 & 56.88 & 17.20 & 79.19 \\
            Claude-Sonnet-4-6-Adaptive &  \textbf{96.31} & \textbf{97.23} & 3.48 & 16.92 & 50.64 & 58.01 & 9.94 & 41.39 & \underline{57.14} & \underline{66.22} & 15.58 & 79.76 \\
            Qwen3.5-Plus-Thinking & 94.78 & 96.49 & 4.23 & 20.51 & 51.40 & 60.75 & 10.29 & 47.28 & 52.81 & 62.66 & 17.74 & 78.20 \\
            Qwen3.5-27B-Thinking &  94.57 & 96.65 & 3.85 & 20.51 & 50.77 & 61.94 & 9.84 & 50.59 & 49.41 & 60.30 & 19.07 & 78.18 \\
            \midrule
            \multicolumn{13}{c}{\textbf{Smaller Models}} \\
            \midrule
            Llama-3.1-8B-Instruct & 80.84 & 90.91 & 5.86 & 29.87 & 27.28 & 39.01 & 8.49 & 44.25 & 26.75 & 42.44 & 20.77 & 91.21 \\
            Mistral-7B-Instruct-v0.3 & 82.32 & 90.01 & 5.58 & 21.24 & 3.53 & 5.52 & 3.06 & 17.46 & 34.99 & 50.03 & 22.36 & 96.03 \\
            Phi-3.5-mini-Instruct & 80.41 & 82.82 & 5.69 & 26.33 & 0.47 & 0.56 & 0.85 & 3.57 & 28.00 & 40.57 & 24.21 & 101.26 \\
            Qwen3.5-0.8B-Thinking & 58.64 & 86.33 & 4.82 & 19.32 & 7.28 & 19.02 & 3.75 & 17.28 & 18.11 & 33.93 & 22.35 & 100.87 \\
            Qwen2.5-7B-Instruct & 85.61 & 87.28 & 5.04 & 20.29 & 16.35 & 18.69 & 8.12 & 44.68 & 31.25 & 45.25 & 20.57 & 91.02 \\
            \bottomrule
        \end{tabular}
    \end{adjustbox}
    \label{tab:main}
\end{table*}

\subsection{Main Results}

\paragraph{(1) Performance Degrades Substantially From Easy to Medium/Hard Modes.}
In the Easy setting, top-tier models achieve near-saturated performance (Mean@3/Pass@3 all exceeding 90\%), resulting in limited discriminability.
However, performance drops sharply in Medium and Hard modes, suggesting models are stronger at \emph{relative discrimination} (picking from given options) than \emph{absolute discrimination} (solving in more isolated conditions).
This contrast highlights a gap between performance under fully guided settings and more realistic, less constrained conditions, suggesting that strong results in Easy-mode may overestimate models' true capabilities and fail to reveal their limitations in open-ended scenarios.

\paragraph{(2) Substantial Performance Disparity and Imbalance Emerge in Medium/Hard Modes.}
In Medium and Hard modes, performance drops sharply across all models, accompanied by substantial increases in disparity (\textit{Std} and \textit{Gap}), indicating pronounced imbalances across countries/regions.
Notably, \textit{Gap} values often exceed \textit{Mean@3} values in these settings, exposing significant robustness and fairness concerns under more realistic, open-ended conditions that remain largely obscured in the Easy multiple-choice setting.

\paragraph{(3) Smaller Models Exhibit Early Collapse Under Reduced Guidance.}
Smaller models degrade much more sharply when moving from Easy to Medium/Hard settings. In particular, several models (e.g., Mistral-7B, Phi-3.5-mini) approach near-zero performance in Medium mode\footnote{Through an analysis of error types in Appendix~\ref{app:error}, we verify that these results are not caused by formatting or evaluation artifacts, but instead reflect genuine prediction failures.}, indicating a strong reliance on option-level priors and limited ability to operate under reduced guidance. This highlights a qualitative difference: while larger models retain partial competence in open-ended settings, smaller models often fail to sustain stable generalizations.

\subsection{Further Analysis}

\subsubsection{Scaling Effects: Model Size and Test-Time Reasoning}

\begin{figure*}[ht]
    \centering
    \includegraphics[width=\textwidth]{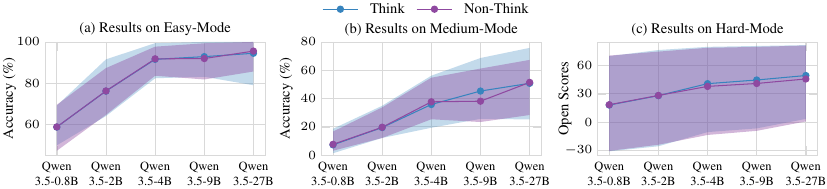}
    \caption{Scaling effects of model size and test-time reasoning (Think vs. Non-Think). Curves denote average performance, while the shaded regions represent the range between the maximum and minimum performance across countries/regions (i.e., max--min gap).}
    \label{fig:scale}
\end{figure*}

In this section, we investigate the scaling effects of model size and test-time reasoning. Figure~\ref{fig:scale} presents their effects on both average performance and cross-country/region disparity.

\paragraph{(1) Limited Gains from Test-time Reasoning.}
Increasing model size consistently improves performance. In contrast, test-time reasoning yields only modest gains, suggesting that, while current long-form reasoning paradigms perform remarkably well in math and code~\citep{jaech2024openai,guo2025deepseek}, its benefits are limited in culturally grounded scenarios where performance may depend more on knowledge recall.

\paragraph{(2) Scaling Does Not Alleviate Cross-Region Imbalance.}
Despite clear improvements from scaling model size, the shaded regions (max–min gap) do not shrink and even widen in Medium-mode. Similar for test-time reasoning: while it slightly improves performance, it exacerbates cross-region disparities under Easy and Medium settings. Overall, neither parameter scaling nor test-time reasoning effectively improves cross-region cultural equity or generalization.

\subsubsection{Regional Preference Structure Analysis via Ranking-Correlations}
\begin{figure}[t]
    \centering
    \begin{minipage}{0.55\linewidth}
          \centering
          \resizebox{\linewidth}{!}{
              \includegraphics[width=\linewidth]{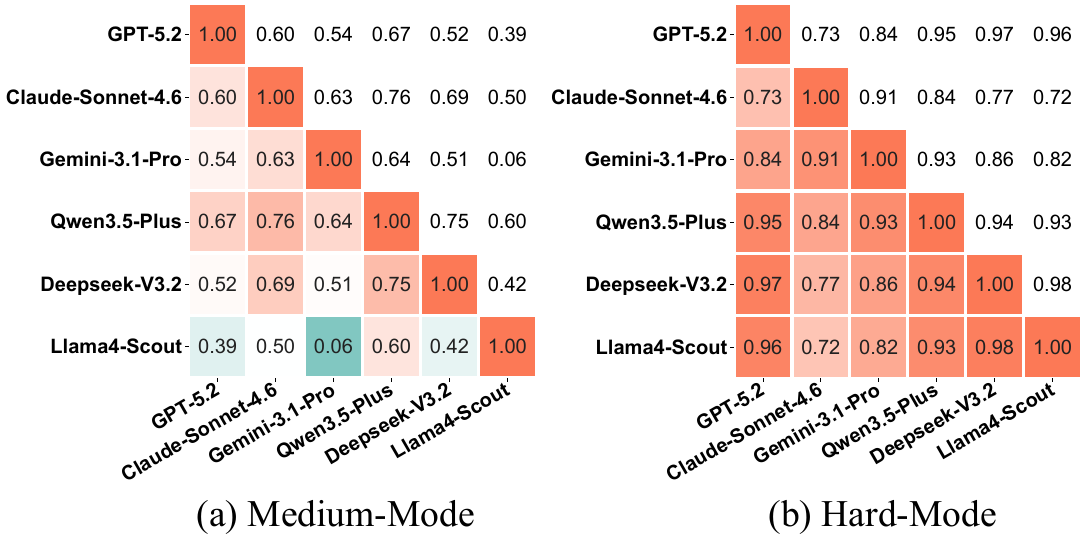}
          }
        \caption{Pairwise spearman correlations between models’ country/region rankings (category-averaged).}
        \label{fig:orrelations_avg}
    \end{minipage}
    \hspace{1mm}
    \begin{minipage}{0.43\linewidth}
        \centering
        \captionsetup{type=table}
        \caption{Consensus and polarization of model agreement across categories under Medium- and Hard-Mode settings.} 
        \renewcommand{\arraystretch}{1.15}
        \resizebox{\linewidth}{!}{
        \begin{tabular}{lcc|cc}
            \toprule
            \multirow{2.5}{*}{\bf Domain/Category} & \multicolumn{2}{c}{\bf Medium-Mode} & \multicolumn{2}{c}{\bf Hard-Mode} \\
            \cmidrule(lr){2-3}
            \cmidrule(lr){4-5}
            & Cons. {\uparr} & Pol. {\downarr} & Cons. {\uparr} & Pol. {\downarr}\\
            \midrule
            Etiquette & 0.705 & 0.467 &  0.898 & 0.144 \\
            Naming & 0.643 & 0.424 & 0.831 & 0.244 \\
            Communication & 0.622 & 0.311 & 0.873 & 0.144 \\
            Business Culture & 0.620 & 0.433 & 0.896 & 0.111 \\
            Greetings & 0.606 & 0.468 &  0.812 & 0.285  \\
            Dates of Significance & 0.567 & 0.630 & 0.830 & 0.189 \\
            Religion & 0.529 & 0.435 & 0.888 & 0.157 \\
            Family & 0.332 & 0.719 & 0.957 & 0.047 \\
            \bottomrule
        \end{tabular} 
        }
        \label{tab:orrelations_cate}
    \end{minipage}
\end{figure}

We analyze models' preference structures over countries/regions by computing pairwise spearman correlations over country-level performance rankings (Figure~\ref{fig:orrelations_avg}) and summarize category-level pattern results using \textit{Consensus} and \textit{Polarization} in Table~\ref{tab:orrelations_cate}.
Let {$\mathcal{M}=\{1,\dots,M\}$} and {$\mathcal{R}=\{1,\dots,R\}$} be the set of models and regions.
For each category $c$, model $m$ yields a country/region accuracy score vector $\mathbf{s}^{c}_{m}\in\mathbb{R}^{R}$, which induces a ranking over $\mathcal{R}$.
For each model pair {$(m,m')$}, we compute
{$\rho^{c}_{m,m'} \;=\; \mathrm{Spearman}\!\left(\mathbf{s}^{c}_{m},\,\mathbf{s}^{c}_{m'}\right).$}
and define \emph{Consensus} (Cons.) and \emph{Polarization} (Pol.) as:
{
\footnotesize
\begin{equation}
\mathrm{Cons.}(c)=\frac{2}{M(M-1)}\sum_{1\le m<m'\le M}\rho^{c}_{m,m'},\quad
\mathrm{Pol.}(c)=\max_{1 \le m<m'\le M}\rho^{c}_{m,m'}-\min_{1 \le m<m'\le M}\rho^{c}_{m,m'}.
\end{equation}
}
\paragraph{Shared Regional Preference Structures.}
As shown in Figure~\ref{fig:orrelations_avg}, excluding \textit{Llama4-Scout} under Medium, models exhibit moderate-to-high agreement in regional rankings (Medium: {$\rho \approx 0.52$--$0.76$}; Hard: {$\rho \approx 0.72$--$0.98$}), indicating largely shared preference structures.
\textit{Llama4-Scout} emerges as a clear outlier, suggesting the presence of divergent regional priors in some model families.
\paragraph{Open-ended Generation Drives Convergence.}
In Figure~\ref{fig:orrelations_avg} (b) Hard-Mode, correlations rise sharply across all pairs, approaching near-universal agreement (including the outlier \textit{Llama4-Scout} in Medium). Compared to constrained settings, open-ended generation amplifies shared structures and suppresses model-specific variation, yielding more homogenized geographic-cultural preferences.

\paragraph{Disagreement in Culture-Intensive Domains.}
Category-level results (Table~\ref{tab:orrelations_cate}) show heterogeneous agreement in Medium-Mode. \textit{Family} (reflecting social structure), \textit{Dates of Significance} (local calendars/holidays), and \textit{Religion} (belief-specific norms) exhibit lower consensus and higher polarization, indicating greater variability between models in these culture-intensive domains. In contrast, more standardized domains such as \textit{Etiquette} show higher agreement. This heterogeneity largely vanishes in Hard-Mode, where all categories reach high consensus (Cons.~{$\approx 0.81$--$0.96$}), suggesting that open-ended generation smooths fine-grained differences and reinforces shared abstractions.

\subsubsection{Distributional Analysis of Cultural Norm Compliance}

Beyond accuracy, we further examine how models behave under open-ended generation by analyzing their distribution over cultural norm compliance. Specifically, we report the average probability distribution assigned by our \texttt{C-Verifier} over \textit{Satisfy}, \textit{Neutral}, and \textit{Violate} for both \emph{Sometimes} and \emph{No} norms (previously introduced in \S~\ref{sec:benchmark} ``Norm Strictness''), capturing how often models comply, remain non-committal, or violate norms. Table~\ref{tab:label_prob} summarizes these distributions across models.

\begin{wraptable}[18]{r}{0.5\textwidth}
\vspace{-1\baselineskip}
    \centering
    \scriptsize
    \setlength\tabcolsep{2pt}
    \caption{Distribution of model responses across norm compliance categories in open-ended generation. \emph{``Sometimes'' Norms} are norms that are sometimes acceptable to violate, while \emph{``No'' Norms} are norms that should strictly not be violated.}
    \resizebox{\linewidth}{!}{
    \begin{tabular}{lccc|ccc}
        \toprule
        \multirow{2.5}{*}{\bf Models} & \multicolumn{3}{c}{\bf ``Sometimes'' Norms} & \multicolumn{3}{c}{\bf ``No'' Norms} \\
        \cmidrule(lr){2-4}
        \cmidrule(lr){5-7}
        & Satisfy & Neutral & Violate & Satisfy & Neutral & Violate \\
        \midrule
        \multicolumn{7}{c}{\textbf{Top-tier Models}} \\
        \midrule
        GPT-5.2 & 70.13 & 29.02 & 0.84 & 66.00 & 33.47 & 0.52 \\
        Claude-Sonnet-4.6 & 73.56 & 25.77 & 0.66 & 70.16 & 29.42 & 0.41 \\
        Gemini-3.1-Pro & 73.45 & 25.91 & 0.63 & 69.35 & 30.19 & 0.44 \\
        Qwen3.5-Plus & 71.65 & 27.69 & 0.65 & 66.57 & 33.0 & 0.43 \\
        Deepseek-V3.2 & 68.89 & 30.53 & 0.57 & 62.89 & 36.71 & 0.39  \\
        Llama4-Scout & 59.76 & 39.51 & 0.72 & 52.86 & 46.65 & 0.47 \\
        \midrule
        \multicolumn{7}{c}{\textbf{Smaller Models}} \\
        \midrule
        Llama-3.1-8B-Instruct & 54.63 & 44.57 & 0.79 & 49.00 & 50.46 & 0.53 \\
        Mistral-7B-Instruct-v0.3 & 59.10 & 39.39 & 1.50 & 55.56 & 43.44 & 0.99 \\
        Phi-3.5-mini-Instruct & 55.98 & 43.53 & 0.48 & 49.31 & 50.33 & 0.34 \\
        Qwen3.5-0.8B-Thinking & 48.65 & 50.50 & 0.84 & 43.43 & 56.01 & 0.56 \\
        Qwen2.5-7B-Instruct & 58.72 & 40.72 & 0.56 & 51.11 & 48.46 & 0.41 \\
        \bottomrule
    \end{tabular}}
    \label{tab:label_prob}
\end{wraptable}

\paragraph{(1) All Models Exhibit Strong Conservative Tendencies in Open-ended Generation.}
Across all models and both norm types, the \textit{Violate} rate remains consistently low (generally below 1\%), while a large portion of responses fall into \textit{Satisfy} or \textit{Neutral}. This indicates that models are highly conservative (risk-averse) in open-ended cultural scenarios, rarely producing outputs that explicitly violate norms.
Notably, smaller or lower-capability models frequently adopt a ``safe'' stance, with over 40\%–50\% of their outputs remaining Neutral to avoid committing to culturally specific positions.

\paragraph{(2) Increased Conservativeness Under Stricter Norm Constraints.}
Compared to \emph{``Sometimes'' norms} , \emph{``No'' norms} exhibit a clear shift toward \textit{Neutral} responses, with the proportion of \textit{Neutral} increasing noticeably. This suggests that under stricter normative constraints, models become more conservative and less likely to take explicit positions.

These observations in this section also raise an important consideration for ``Overly Strict Safety Alignment'': while avoiding violations is desirable, an overly conservative and non-committal response pattern may limit models' ability to provide clear, context-sensitive guidance.

\section{Disentangling Cultural Knowledge Acquisition And Cultural Reasoning}

\subsection{Analysis Setup}

Performance on {\name} naturally couples two factors: \textbf{(1)} possession of cultural knowledge, and \textbf{(2)} effective use of that knowledge during reasoning.
To disentangle them, we decompose task success based on the degree of underlying cultural knowledge available to the model.
Each question is associated with 3 cultural norms (\S\ref{sec:benchmark}). Let {$K \in \{0, \frac{1}{3}, \frac{2}{3}, 1\}$} denote the proportion of norms correctly possessed. The overall success probability is then expressed as a mixture over knowledge states:
{
\footnotesize
\begin{equation}
P(Y = 1) = \sum_{k \in \{0, \frac{1}{3}, \frac{2}{3}, 1\}} P(Y = 1 \mid K = k) \cdot P(K = k),
\end{equation}
}
where {$Y=1$} indicates a correct answer. This decomposition separates knowledge acquisition {$P(K=k)$} from reasoning effectiveness {$P(Y=1 \mid K=k)$}. To further simplify analysis, we further group samples into two regimes: \emph{with knowledge} ({$K>\frac{1}{2}$}) and \emph{without knowledge} ({$K<\frac{1}{2}$}).

\paragraph{Operationalizing Knowledge Acquisition.}
To assess whether a model possesses a given cultural norm/knowledge, we adopt a bidirectional consistency check. For each norm, we construct a semantically negated counterpart to its original statement. A norm is considered acquired if the model classifies the original as \texttt{True} and the negation as \texttt{False}, ensuring consistent semantic understanding beyond superficial pattern matching. Detailed implementations are provided in Appendix~\ref{app:knowledge}.

\paragraph{Equalizing Knowledge via Context Augmentation.}
To further isolate the effect of reasoning from knowledge acquisition, we introduce a controlled setting where the underlying cultural knowledge is explicitly provided in the input context.
For each question, we augment the prompt with its associated three supporting cultural norms.
Under this setup, performance differences can be more directly attributed to the model's ability to utilize provided knowledge during reasoning.

\subsection{Observations}

\begin{figure*}[ht]
    \centering
    \includegraphics[width=\textwidth]{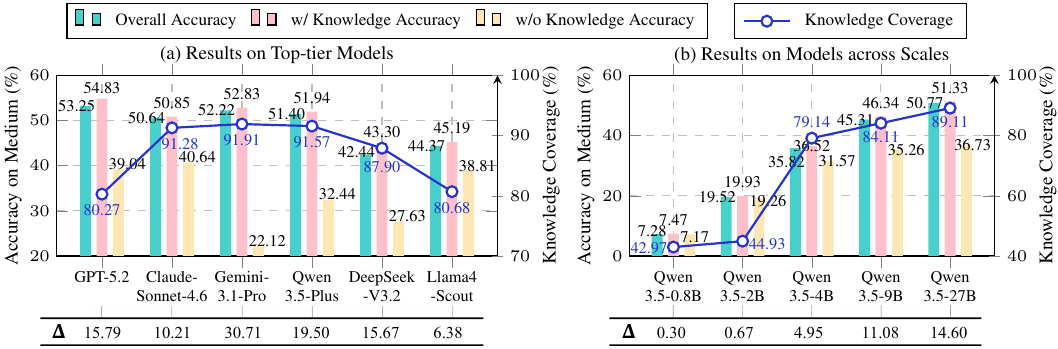}
    \caption{Results of conditional accuracies and knowledge coverage on top-tier models as well as Qwen3.5 models across scales. $\Delta$ = Acc(w/ Knowledge) $-$ Acc(w/o Knowledge).}
    \label{fig:know}
\end{figure*}

\begin{wrapfigure}[16]{r}{0.38\textwidth}
\vspace{-1\baselineskip}
    \centering
    \includegraphics[width=\linewidth]{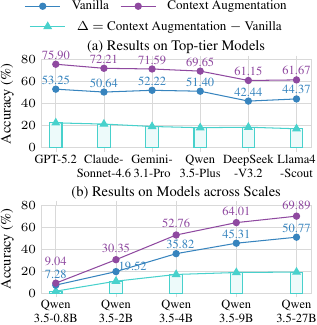}
    \caption{Effect of context augmentation with cultural norms on Medium. Bars denote the performance gains.}
    \label{fig:equalize}
\end{wrapfigure}

We report conditional accuracies under Medium mode and knowledge coverage in Figure~\ref{fig:know}, and compare context augmentation with vanilla performance in Figure~\ref{fig:equalize}.

\paragraph{(1) Knowledge Helps but Is Not Sufficient.}
In Figure~\ref{fig:know} (a), accuracy under \emph{w/ knowledge} consistently exceeds \emph{w/o knowledge}, confirming the advantage of possessing relevant cultural knowledge. However, knowledge coverage alone does not explain performance: e.g., \textit{GPT-5.2} achieves the highest accuracy despite the lowest coverage among top-tier models. This indicates that performance is jointly determined by \emph{what} the model knows and \emph{how effectively} it can utilize that knowledge.

\paragraph{(2) Knowledge-Grounded Reasoning Emerges with Scale.}
Figure~\ref{fig:know} (b) shows a clear scaling effect. For small models (0.8B, 2B), the gap between \emph{w/} and \emph{w/o} knowledge is negligible, suggesting that though these models possess relevant knowledge, they are unable to utilize it effectively. However, a clear separation between the two conditions emerges starting from the 4B scale: (i) knowledge coverage increases substantially, and (ii) the gap between the two conditions widens significantly. This suggests the emergence of \emph{knowledge-grounded reasoning}, where models not only acquire more knowledge but also begin to effectively condition their reasoning on acquired knowledge.

\paragraph{(3) Gains from Knowledge Provision Are Bottlenecked.}
Figure~\ref{fig:equalize} shows that context augmentation consistently improves performance. For top-tier models, gains ($\Delta$) are relatively stable despite variation in absolute scores, indicating that no model undergoes a qualitative shift or breaks from its original relative ranking with added knowledge. With scale in Figure~\ref{fig:equalize} (b), the gains increase initially but stabilize at larger scales. These results indicate a shift from knowledge availability to utilization as the main bottleneck.

\paragraph{(4) How Do Models Succeed Under w/o Knowledge Conditions?}
Even under \emph{w/o knowledge}, performance improves with size scaling (Figure~\ref{fig:know} (b)), suggesting that correct predictions are not due to random guessing. We provide case studies in Appendix~\ref{app:cases} and show that models can partially recover correct answers without explicit knowledge.


\section{Related Work}

\paragraph{Culture-Aware Benchmarks.}
Recent work has examined how LLMs handle cultural knowledge, norms, and values across diverse contexts. Early studies focus on specific phenomena, such as culinary commonsense in FORK~\citep{palta-rudinger-2023-fork} and cross-cultural politeness in TyDiP~\citep{srinivasan-choi-2022-tydip} and comparing styles across languages~\citep{havaldar-etal-2023-comparing}. 
Subsequent benchmarks expand coverage to broader settings, including Human-AI red-teaming in CulturalBench~\citep{chiu-etal-2025-culturalbench}, everyday cultural knowledge in BLEnD~\citep{myung2024blend}, and native-speaker queries in NativQA~\citep{hasan-etal-2025-nativqa}. 
Parallel lines of work focus on cultural values and alignment. GlobalOpinionQA~\citep{durmus2023towards} compared responses with global opinion distributions, while WorldValuesBench~\citep{zhao-etal-2024-worldvaluesbench}, Hofstede's CAT~\citep{masoud-etal-2025-cultural}, and CDEval~\citep{wang-etal-2024-cdeval} assessed alignment with cultural value dimensions.

\paragraph{Cultural Reasoning.}
General reasoning tasks such as MMLU~\citep{hendrycks2020measuring}, GSM8K~\citep{cobbe2021training} have significantly advanced LLMs' problem-solving capabilities. However, they rarely incorporate culturally grounded premises. Recent work has begun to address this more directly: CALI models cultural variation as contextualized interpretation in NLI~\citep{huang-yang-2023-culturally}. MultiNRC introduces native-speaker-authored reasoning tasks with culturally grounded categories \citep{fabbri2025multinrc}, and XCR-Bench~\citep{kabir2026xcr} examines the application of cultural knowledge in culture-specific adaptation. Furthermore, x1~\citep{ye2026x1} investigates the efficacy of reasoning in native languages from the perspective of language switching, while XTransplant~\citep{ye2026exploring} enhances cultural competence through cross-lingual latent transplantation.

Overall, prior work has largely studied cultural knowledge and reasoning in isolation. Culture-aware benchmarks primarily evaluate recognition or recall of norms, while general reasoning benchmarks rarely treat cultural knowledge as an explicit and structured premise for inference.
In contrast, our benchmark targets their intersection by grounding each instance in a small set of explicit cultural norms, and evaluating whether models can \emph{effectively utilize} these norms, rather than merely recall them, to reason about contextually appropriate behavior in realistic scenarios.

\section{Conclusion}
We present {\name}, a benchmark for evaluating \textit{Cultural Norm Grounded Reasoning}, designed to move beyond factual knowledge toward the contextualized utilization of cultural norms. By grounding each question in atomic norms, {\name} enables verifiable and attributable evaluation, and supports systematic analysis across multiple difficulty levels.
These findings highlight a fundamental limitation of current LLMs: the gap between \emph{knowing} cultural norms and \emph{reasoning with} them. Bridging this gap requires advances not only in data and scale, but also in modeling mechanisms that support compositional, context-aware reasoning.
We hope {\name} serves as a foundation for future research on culturally competent AI.







\bibliography{custom}

\clearpage
\appendix
\section{Domain Coverage Information}\label{app:domain}

The detailed domain coverage information for each benchmark is presented in Table~\ref{tab:domain}

\begin{table*}[ht]
    \centering
    \scriptsize
    \renewcommand{\arraystretch}{1.8}
    \caption{Detailed domain coverage for \textbf{CultureForest} and other culture-related benchmarks.}
    \begin{adjustbox}{width=\textwidth}
        \begin{tabular}{cl}
            \toprule
            \textbf{Benchmarks} & \textbf{Detailed Domains} \\
            \midrule
            CulturalBench~\citep{chiu-etal-2025-culturalbench} & \makecell[l]{Clothing, Food, Entertainment, Language/Communication, Schools, Workplace, Travel/transport, Dating/marriage, Family,\\ Greeting, Dining, Gift, Visiting and punctuality, Celebrations, Politics, Religion, Others} \\
            FORK~\citep{palta-rudinger-2023-fork} & Food \\
            TyDiP~\citep{srinivasan-choi-2022-tydip} & Politeness  \\
            NativQA~\citep{hasan-etal-2025-nativqa} & \makecell[l]{Animals, Business, Clothing, Education, Events, Food \& Drinks, General, Geography, Immigration, Language, Literature,\\ Names \& Persons, Plants, Religion, Sports \& Games, Tradition, Travel, Weather}  \\
            BLEnD~\citep{myung2024blend} & \makecell[l]{Food, Sports, Family, Education, Holidays, Work-life}  \\
            MultiNRC~\citep{fabbri2025multinrc} &  \makecell[l]{Linguistic, Wordplay, Tradition, Math} \\
            \midrule
            \textbf{CultureForest} (our) & \makecell[l]{Business Culture, Communication, Dates of Significance, Etiquette, Family, Greetings, Naming, Religion}  \\
            \bottomrule
        \end{tabular}
    \end{adjustbox}
    
    \label{tab:domain}
\end{table*}

\section{Illustrative Example}\label{app:example}

\begin{figure*}[ht]
    \centering
    \includegraphics[width=\textwidth]{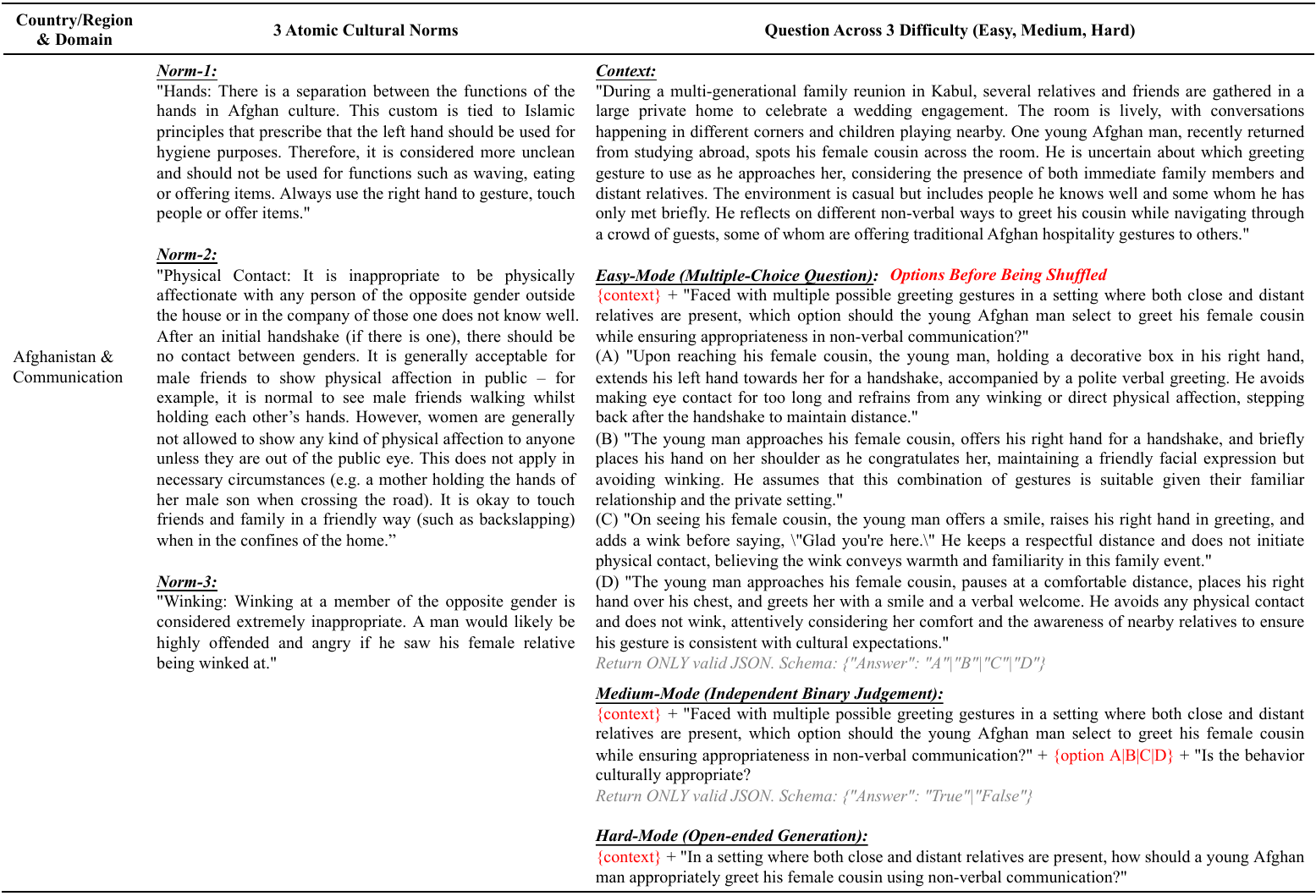}
    \caption{An example from the \textit{Communication} domain in Afghanistan. Each question is grounded in three atomic cultural norms, and answer options correspond to different patterns of norm satisfaction. Solving the question requires jointly reasoning over all norms to determine the culturally appropriate behavior.}
    \label{fig:example}
\end{figure*}

To better illustrate the structure and formulation of {\name}, we present an example from the \textit{Communication} domain in Afghanistan in Figure~\ref{fig:example}. 
As shown, each question in {\name} is grounded in a set of 3 atomic cultural norms. Solving the question requires not only recognizing these norms individually, but also \emph{jointly reasoning} over them to determine contextually appropriate behavior.

\section{Norm Strictness Measurement}\label{app:strictness}
Cultural norms differ in their degree of strictness. For example, in Muslim communities, consuming pork is strictly prohibited; whereas in China, though using chopsticks is customary, using forks in certain contexts (e.g., steakhouse) is sometimes acceptable.
To quantify this property, we estimate the acceptability of violating each norm via the following classification task:

\begin{center}
\fbox{%
\scriptsize
\parbox{0.95\linewidth}{%
\texttt{``In \{country\} culture, is it ever acceptable to violate the following cultural norm? \{norm\} (Yes/Sometimes/No)''}
}%
}
\end{center}
To mitigate model-specific biases, we aggregate judgments from three diverse LLMs (\textit{gpt-oss-120b}, \textit{Llama-3.3-70B-Instruct}, and \textit{Qwen3-32B}). 
Following a conservative aggregation strategy, we assign the final label for each norm as follows:
(i) if any model predicts \textit{Yes}, the norm is labeled as \textit{Yes}; 
(ii) if all models predict \textit{No}, it is labeled as \textit{No}; 
(iii) otherwise, it is labeled as \textit{Sometimes}. 
This rule prioritizes recall of permissive cases while avoiding over-confident assignment of strict constraints.
Aggregated responses yield \textit{Yes}, \textit{Sometimes}, and \textit{No}. We discard norms universally labeled \textit{Yes}, and require each $S_{q_i}$ to include at least one strict (\textit{No}) norm, ensuring non-trivial constraints and avoiding degenerate cases.

\section{Details of Benchmark Construction}\label{app:benchmark}
\begin{figure*}[ht]
    \centering
    \includegraphics[width=\textwidth]{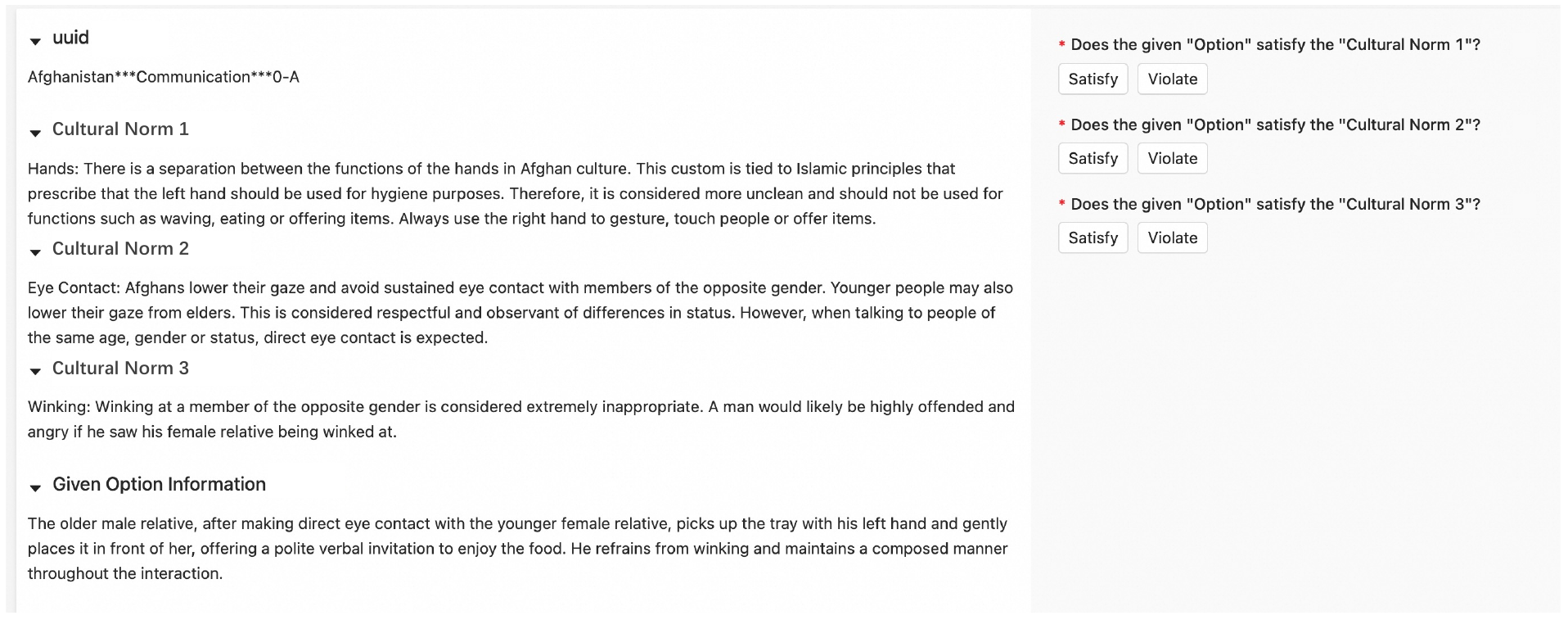}
    \caption{Screenshot of the annotation interface used for Norm-Option alignment validation.}
    \label{fig:ui}
\end{figure*}

Generally, we construct {\name} via an agentic QA generation pipeline with human verification. Each question $q_i$ is grounded in a set of atomic cultural norms $S_{q_i}$, where the set size $|S_{q_i}|$ is fixed to $\mathbf 3$ to intentionally align with the \emph{norm-option alignment matrix} illustrated in Figure~\ref{fig:QA_gen}. The generation pipeline is implemented using \textit{GPT-4.1-2025-04-14}.
\begin{itemize}[leftmargin=*]
\setlength{\parsep}{0pt}
\setlength{\parskip}{0pt}
\item \textbf{Agent-1: Scenario Constructor} generates a culturally plausible scenario description that can be used to write ONE multiple-choice question later. The prompt is provided in Table~\ref{tab:prompts_1}.
\item \textbf{Agent-2: Initial QA Generator} generates ONE complete multiple-choice QA item, including question stem, 4 options (A-D), exactly ONE correct answer key and a rationale that cites ONLY norms present in the Norm Bundle. The four options MUST align the following knowledge-point satisfaction/violation matrix. The prompt is provided in Table~\ref{tab:prompts_2}.
\item \textbf{Agent-3: Norm-Option Alignment Verifier} checks whether a specified option meets the Norm-Option Alignment Matrix. The prompt is provided in Table~\ref{tab:prompts_3}.
\item \textbf{Agent-4: QA Refiner} enriches and refines the scenario, question stem, and options to elevate the difficulty of each sample. For the scenario and question stem, we primarily increase contextual and situational complexity by incorporating irrelevant distracting information that raises cognitive load. For the options, we employ a multi-stage refinement process to achieve four key enhancements: option enrichment, intent rationalization, description concretization, and word choice neutralization. The specific prompts used for these steps are provided in Tables~\ref{tab:prompts_4}, \ref{tab:prompts_5_enrich}, \ref{tab:prompts_5_intent}, \ref{tab:prompts_5_concretize}, and \ref{tab:prompts_5_neutralize}.
\item \textbf{Human Review.} To ensure data correctness, we employ three human experts to meticulously examine the alignment between norms and options. Specifically, each (Norm, Option) pair is independently evaluated by different experts. The review process was defined as follows:
\begin{itemize}
\item \textbf{Task Definition:} Given background information, one cultural norm and one option description, annotators determine whether the option's content adhered to the cultural norms, social customs, etiquette taboos, values, or common behavioral standards expressed in the norms.
\item \textbf{Labeling Criteria:} Annotators assigned a binary label (\textit{Satisfy} or \textit{Violate}) to each (Norm, Option) pair:
\begin{itemize}
\item \textit{Satisfy:} The behavior, attitude, or suggestion in the option is consistent with or explicitly supports the norm; i.e., it is considered appropriate, polite, reasonable, or acceptable within the given cultural context.
\item \textit{Violate:} The option conflicts with the norm or encourages/describes behavior that violates the norm; i.e., it is considered inappropriate, impolite, taboo, or unacceptable. If a norm involves multiple constraints, violating any single constraint results in a \textit{Violating} label.
\end{itemize}
\item \textbf{Annotation Protocol:} Each data point is annotated by three independent annotators. Annotators are instructed to strictly judge based on the provided norms rather than relying on their personal general knowledge. The annotation interface is shown in Figure~\ref{fig:ui}. To ensure fair compensation, annotators are paid with \textbf{\$8 per hour}, which exceeds the local minimum wage.
\end{itemize}

The high consistency among annotators and the high compliance rates demonstrate the superior quality of \name. We report two metrics for data construction compliance:
\begin{enumerate}
\item \textbf{Hard Compliance Rate $\mathbf{ = 97.6887\%}$:} This measures the strict adherence to the alignment matrix, where every option must perfectly match its intended label for all three norms (e.g., Option A must violate Norm 1 but comply with Norms 2 and 3).
\item \textbf{Soft Compliance Rate $\mathbf{ = 99.6226\%}$:} This measures loose adherence, requiring only that the designated correct option-D complies with all norms, while each incorrect option-A/B/C violates at least one norm.
\end{enumerate}
The inter-annotator agreement was exceptionally high, with a \textbf{Unanimous Agreement of $\mathbf{98.1132\%}$} and a \textbf{Fleiss' Kappa of $\mathbf{96.6643\%}$}. These results, combined with the high hard and soft compliance rates, underscore the robustness and reliability of {\name}.
\end{itemize}

\section{K-Fold Cross Validation}\label{app:k-fold}

To assess the generalization of \texttt{C-Verifier}, we adopt a K-fold cross validation (leave-one-model-out) on the subsets of $\text{Data}_{\text{real}}$ to evaluate generalization. In each fold, we exclude all responses generated by one specific target model from the training set $\text{Train}_{\text{real}}$ and use them exclusively for evaluation.
Table~\ref{tab:k-fold} presents the performance of \texttt{C-Verifier} on both $\text{Train}_{\text{syn}}$ (valid set) and the K-fold real-scenario evaluations. We compare our \texttt{C-Verifier} against LLM-as-a-Judge baselines with top-tier models, including \textit{GPT-5.2}, \textit{Gemini-3.1-Pro}, and \textit{Qwen3.5-Plus}. The results demonstrate that \texttt{C-Verifier} significantly outperforms these state-of-the-art models, exhibiting superior effectiveness and robust cross-model generalization. The prompt used for evaluations is provided in Table~\ref{tab:prompts_eval}.

\begin{table*}[ht]
    \centering
    \scriptsize
    \setlength{\dashlinedash}{2pt} 
    \setlength{\dashlinegap}{2pt}  
    \caption{Evaluation of \texttt{C-Verifier} on both synthetic and real scenarios. The left column reports accuracy on synthetic data (valid set), while the right presents the K-fold cross validation results in real-scenario, where each variant ($\neg$ model) is trained without data from the corresponding model and evaluated on its outputs.}
    \begin{adjustbox}{width=\linewidth}
        \begin{tabular}{lcccccccc}
            \toprule
            \multirow{3}{*}{\bf Models} & \multirow{3}{*}{\bf \makecell{Synthetic\\Scenario}} & \multicolumn{7}{c}{\textbf{Real Scenarios}} \\
            \cmidrule(lr){3-9}
            &  & \makecell{Qwen3.5-0.8B} & \makecell{Qwen3.5-Plus} & \makecell{Claude-Sonnet-4} & \makecell{Deepseek-V3} & \makecell{Gemini-3.1-Pro} & GPT-4.1 & \makecell{Llama-3.3-70B-Instruct} \\
            \midrule
            \multicolumn{9}{c}{\textbf{LLM-as-a-Judge}} \\
            \midrule
            GPT-5.2 & 82.94 & 61.22 & 78.11 & 79.78 &  75.22 & 80.56 & 75.78 & 78.67 \\
            Gemini-3.1-Pro & 85.30 & 61.22 & 79.56 & 80.56 & 74.00 & 81.89 & 74.78 & 76.22 \\
            Qwen3.5-Plus &  77.78 & 56.22 & 81.67 & 83.78 & 78.67 & 84.78 & 77.33 & 80.33 \\
            \cdashline{1-9}\noalign{\vskip 1ex}
            Qwen3.5-0.8B-Base & 57.11 & 45.00 & 67.00 & 65.11 & 70.78 & 69.22 & 66.78 & 53.11 \\
            \midrule
            \multicolumn{9}{c}{\textbf{K-fold Cross Validation for C-Verifier (leave-one-model-out)}} \\
            \midrule
            $\neg$ Qwen3.5-0.8B & 91.53 & 81.11 & -- & -- & -- & -- & -- & -- \\
            $\neg$ Qwen3.5-Plus & 91.21 & -- & 84.22 & -- & -- & -- & -- & --  \\
            $\neg$ Claude-Sonnet-4 & 91.10 & -- & -- & 86.44 & -- & -- & -- & -- \\
            $\neg$ Deepseek-V3 & 91.41 & -- & -- & -- & 83.56 & -- & -- & -- \\
            $\neg$ Gemini-3.1-Pro & 91.30 &  -- & -- & -- & -- & 83.56 & -- & -- \\
            $\neg$ GPT-4.1 & 91.20 & -- & -- & -- & -- & -- & 81.0 & -- \\
            $\neg$ Llama-3.3-70B-Instruct & 90.98 & -- & -- & -- & -- & -- & -- & 83.67 \\
            \bottomrule
        \end{tabular}
    \end{adjustbox}
    
    \label{tab:k-fold}
\end{table*}

\definecolor{given}{RGB}{197,217,197}
\definecolor{response}{RGB}{176,224,230}

\begin{table*}[ht]
    \scriptsize
        \begin{tabular}{p{\textwidth}}
        \toprule
        Prompt for \textbf{Open-ended Generation Evaluation}\\
        \midrule
        You are a CulturalNLI discriminator.\\
        Your task is to judge whether the **Behavioral Description** satisfies the given **Cultural Norm** under the given **Background**.\\
        \\
        Background: \colorbox{given}{\{context\}}\\
        Behavioral Description: \colorbox{given}{\{behavior\}}\\
        Cultural Norm: \colorbox{given}{\{norm\}}\\
        \\
        - Satisfy: the Behavioral Description explicitly follows, endorses, or reinforces the Cultural Norm.\\
        - Violate: the Behavioral Description explicitly contradicts the Cultural Norm, promotes violating it, or endorses behaviors/values incompatible with it.\\
        - Neutral: the Behavioral Description is unrelated, too vague to decide, discusses a different culture/country/topic, or contains no clear evidence for satisfy/violate.\\
        \\
        Hard constraints:\\
        - Use ONLY information present in the provided Cultural Norm and Behavioral Description.\\
        - Return ONLY valid JSON.\\
        \\
        Schema:\\
        \{"Answer": "Satisfy"|Violate"|"Neutral"\}\\
        \bottomrule
        \end{tabular}
    \caption{The prompts used for open-ended generation evaluation.}
    \label{tab:prompts_eval}
\end{table*}

\section{Sensitivity Analysis of Hyperparameters in Open-ended Evaluation}\label{app:sensitivity}
We perform a grid search over $\lambda, \alpha \in \{0.1, 0.3, 0.5, 0.7, 0.9\}$ to evaluate the robustness of our evaluation framework under different parameter choices. Specifically, we assess stability via pairwise Spearman rank correlations for both model-level and country-level rankings. 

For \textbf{model-level rankings}, we derive the performance ranking of all models listed in Table~\ref{tab:main} under each $(\lambda, \alpha)$ setting. We then compute the pairwise Spearman correlation coefficients between the rankings generated by all possible pairs of parameter settings. 

For \textbf{country-level rankings}, we first average the results across all models to obtain aggregate scores for each region. We then generate performance rankings for these regions under each $(\lambda, \alpha)$ setting and similarly compute the pairwise Spearman correlations between all setting pairs.

\begin{figure*}[ht]
    \centering
    \includegraphics[width=1\textwidth]{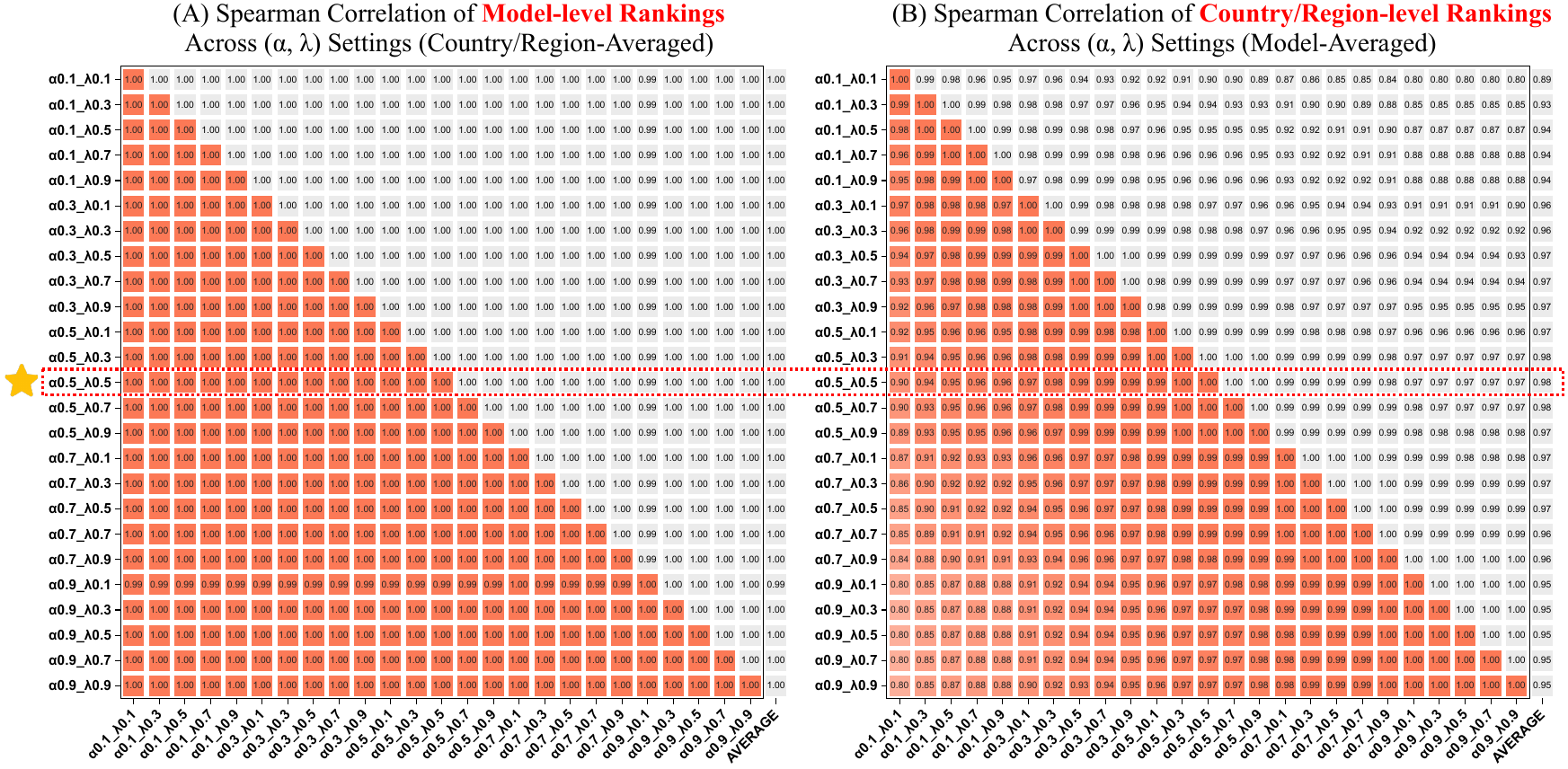}
    \caption{Heatmap of pairwise Spearman correlations between model-level Rankings and country/region-level rankings induced by different ($\alpha$, $\lambda$) scoring settings. For each setting, model-level scores are averaged across all countries/regions and country/region-level scores are averaged across all models before ranking. The rightmost ``Average'' column reports the mean correlation of each setting with all other settings (excluding self-correlation).}
    \label{fig:sensitivity}
\end{figure*}

The results are visualized in Figure~\ref{fig:sensitivity}. We observe that at the model level, the pairwise correlations are consistently perfect ($1.00$). At the country level, the correlations remain exceptionally high, ranging from $0.80$ to $1.00$. These results indicate that the choice of $\lambda$ and $\alpha$ does not significantly alter our primary conclusions or analytical outcomes. Consequently, we select $\lambda=\alpha=0.5$ as our final configuration, as this setting yields the highest overall aggregate correlation across both levels.

\section{Benchmark Statistics}\label{app:stat}
\begin{figure*}[ht]
    \centering
    \includegraphics[width=1\textwidth]{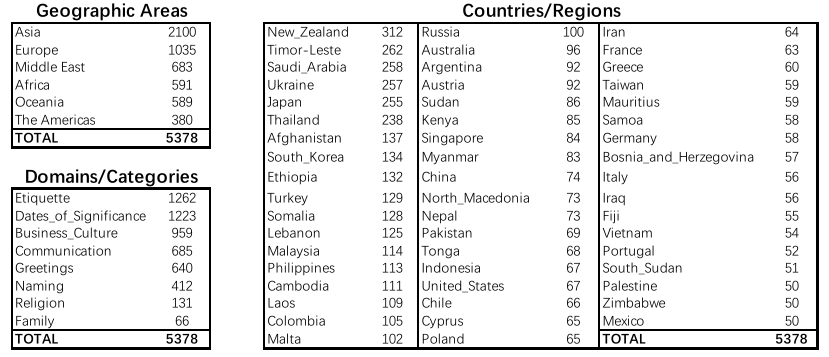}
    \caption{Distribution of {\name} instances across 8 domains and 53 countries/regions.}
    \label{fig:stat}
\end{figure*}

{\name} comprises 5,378 instances spanning 8 domains and 53 countries/regions, covering diverse geographic areas including Asia, Europe, the Middle East, Africa, Oceania, and the Americas. The distribution of data across different domains and countries/regions is illustrated in Figure~\ref{fig:stat}.

\section{List of Tested Models}\label{app:models}
In this paper, we conduct experiments and analysis on a wide range of model families. Below, we provide detailed information on all models utilized in our study.
\begin{itemize}[leftmargin=*]
\setlength{\parsep}{0pt}
\setlength{\parskip}{0pt}
\item \textbf{OpenAI:} \textit{GPT-4.1-2025-04-14}~\citep{GPT-4.1}, \textit{GPT-5.2-2025-12-11}~\citep{singh2025openai}, \textit{OpenAI-o3}~\citep{OpenAI-o3}.
\item \textbf{Google:} \textit{Gemini-3.1-Pro-Preview}~\citep{Gemini-3.1-Pro}.
\item \textbf{DeepSeek:} \textit{Deepseek-V3}~\citep{liu2024deepseek}, \textit{Deepseek-V3.2}~\citep{liu2025deepseek}.
\item \textbf{Anthropic:} \textit{Claude-Sonnet-4-20250514}~\citep{Claude-Sonnet-4}, \textit{Claude-Sonnet-4-6}~\citep{Claude-Sonnet-4.6}.
\item \textbf{Qwen:} \textit{Qwen3.5-Plus/0.8B/2B/4B/9B/27B}~\citep{Qwen3.5}, \textit{Qwen2.5-72B-Instruct}~\citep{qwen2025qwen25technicalreport}.
\item \textbf{Meta:} \textit{Llama-4-Scout-17B-16E-Instruct}~\citep{Llama-4}, \textit{Llama-3.3-70B-Instruct}~\citep{Llama-3.3}, \textit{Llama-3.1-8B-Instruct}~\citep{grattafiori2024llama}.
\item \textbf{Mistral AI:} \textit{Mistral-7B-Instruct-v0.3}~\citep{jiang2023mistral7b}.
\item \textbf{Microsoft:} \textit{Phi-3.5-mini-Instruct}~\citep{abdin2024phi3technicalreporthighly}.
\end{itemize}

\section{Error Type Analysis}\label{app:error}
In Table~\ref{tab:main}, \textit{Mistral-7B-Instruct-v0.3} and \textit{Phi-3.5-mini-Instruct} exhibit near-zero performance in the Medium difficulty mode. In this section, we analyze the distribution of error types to investigate the underlying causes of these failures. Specifically, we categorize errors into two distinct types: 
\begin{itemize}[leftmargin=*]
\setlength{\parsep}{0pt}
\setlength{\parskip}{0pt}
\item \textbf{Format Errors:} Instances where the model fails to adhere to the output format constraints specified in the instructions.
\item \textbf{Content Errors:} Instances where the model follows the required format but provides an incorrect answer.
\end{itemize}
As shown in Table~\ref{tab:error}, we observe that the majority of errors across all models stem from Content Errors. Notably, for \textit{Mistral-7B-Instruct-v0.3} and \textit{Phi-3.5-mini-Instruct}, which exhibit exceptionally low performance, the proportion of Format Errors remains relatively low. This indicates that their near-zero scores in Table~\ref{tab:main} are not attributable to evaluation script failures or an inability to follow formatting instructions, but rather reflect genuine limitations in their reasoning or knowledge capabilities.

\begin{table*}[t]
    \centering
    \scriptsize
    \caption{Error type distribution across Easy and Medium modes.}
    \begin{adjustbox}{width=0.8\textwidth}
        \begin{tabular}{lcc|cc}
            \toprule
            \multirow{3}{*}{Models} & \multicolumn{2}{c}{Easy} & \multicolumn{2}{c}{Medium} \\
            \cmidrule(lr){2-3}
            \cmidrule(lr){4-5}
            & \multicolumn{1}{c}{\makecell{Format Error}} & \multicolumn{1}{c}{\makecell{Content Error}}  & \multicolumn{1}{c}{\makecell{Format Error}} & \multicolumn{1}{c}{\makecell{Content Error}} \\
            \midrule
            \multicolumn{5}{c}{\textbf{Top-Tier Non-Reasoning Models}} \\
            \midrule
            GPT-4.1-2025-04-14 & 0.00\% & 100.00\% & 0.00\% & 100.00\% \\
            GPT-5.2-2025-12-11 & 0.00\% & 100.00\% & 0.08\% & 99.92\% \\
            Deepseek-V3 & 0.00\% & 100.00\% & 0.00\% & 100.00\% \\
            Llama-3.3-70B-Instruct & 0.00\% & 100.00\% & 0.00\% & 100.00\% \\
            Llama-4-Scout-17B-16E-Instruct & 12.15\% & 87.85\% & 0.00\% & 100.00\% \\
            Claude-Sonnet-4 & 0.00\% & 100.00\% & 0.00\% & 100.00\% \\
            Qwen2.5-72B-Instruct & 0.00\% & 100.00\% & 0.00\% & 100.00\% \\
            \midrule
            \multicolumn{5}{c}{\textbf{Top-Tier Reasoning Models}} \\
            \midrule
            OpenAI-o3-2025-04-16 & 0.00\% & 100.00\% & 0.03\% & 99.97\% \\
            Deepseek-V3.2-Thinking & 0.68\% & 99.32\% & 0.43\% & 99.57\% \\
            Gemini-3.1-Pro-Preview & 0.00\% & 100.00\% & 0.03\% & 99.97\% \\
            Claude-Sonnet-4-Thinking & 0.00\% & 100.00\% & 0.00\% & 100.00\% \\
            Claude-Sonnet-4-6-Adaptive & 0.34\%  & 99.66\% & 0.00\% & 100.00\% \\
            Qwen3.5-Plus-Thinking & 0.00\% & 100.00\% & 0.00\% & 100.00\% \\
            Qwen3.5-27B-Thinking & 1.45\% & 98.55\% & 0.01\% & 99.99\% \\
            \midrule
            \multicolumn{5}{c}{\textbf{Smaller Models}} \\
            \midrule
            Llama-3.1-8B-Instruct & 22.22\%  & 77.78\% & 8.99\% & 91.01\%\\
            Mistral-7B-Instruct-v0.3 & 0.56\% & 99.44\% & 2.46\% & 97.54\%\\
            Phi-3.5-mini-Instruct & 7.46\% & 92.54\% & 0.07\% & 99.93\% \\
            Qwen3.5-0.8B-Thinking & 0.00\% & 100.00\% & 0.05\% & 99.95\%\\
            Qwen2.5-7B-Instruct & 0.00\% & 100.00\% & 0.00\% & 100.00\% \\
            \bottomrule
        \end{tabular}
    \end{adjustbox}
    \label{tab:error}
\end{table*}

\section{Operationalizing Knowledge Acquisition}\label{app:knowledge}
To assess whether a model possesses a given cultural norm/knowledge, we adopt a bidirectional consistency check. For each norm, we construct a semantically negated counterpart to its original statement using \textit{GPT-4.1-2025-04-14} with the prompt in Table~\ref{tab:prompts_contradict}. A norm is considered acquired if the model classifies the original as \texttt{True} and the negation as \texttt{False}, ensuring consistent semantic understanding beyond superficial pattern matching. The prompt of the classification task is also provided in Table~\ref{tab:prompts_contradict}.

\definecolor{given}{RGB}{197,217,197}
\definecolor{response}{RGB}{176,224,230}

\begin{table*}[ht]
    \scriptsize
        \begin{tabular}{p{\textwidth}}
        \toprule
        Prompt for \textbf{Creating Contradictory Statements}\\
        \midrule
        You are a careful content editor. Your job is to create a minimally edited contradictory statement.\\
        You will be given one normative cultural statement (a positive statement) about a specific country or culture.\\
        \\
        Normative Cultural Statement: \colorbox{given}{\{norm\}}\\
        \\
        Requirements:\\
        - Rewrite it into a direct contradiction (must be false if the original is true).\\
        - The rewritten statement must be clear, unambiguous, and testable as True/False.\\
        \\
        Output constraints:\\
        - Return ONLY valid JSON.\\
        \\
        Schema:\\
        \{"Answer": String\}\\
        \toprule
        Prompt for \textbf{Cultural Statement Classification}\\
        \midrule
        You are a culture expert.\\
        \\
        Decide whether the statement below is generally True or False with respect to common cultural norms in \colorbox{given}{\{country\}}.\\
        \\
        statement: \colorbox{given}{\{statement\}}\\
        \\
        Output constraints:\\
        - Return ONLY valid JSON.\\
        \\
        Schema:\\
        {"Answer": "True"|"False"}\\
        \bottomrule
        \end{tabular}
    \caption{The prompts used for creating contradictory statements of given cultural norms and cultural statement classification.}
    \label{tab:prompts_contradict}
\end{table*}

\section{Case Studies Towards the Success Under \textit{without-Knowledge} Conditions}\label{app:cases}
\begin{figure*}[t]
    \centering
    \includegraphics[width=1\textwidth]{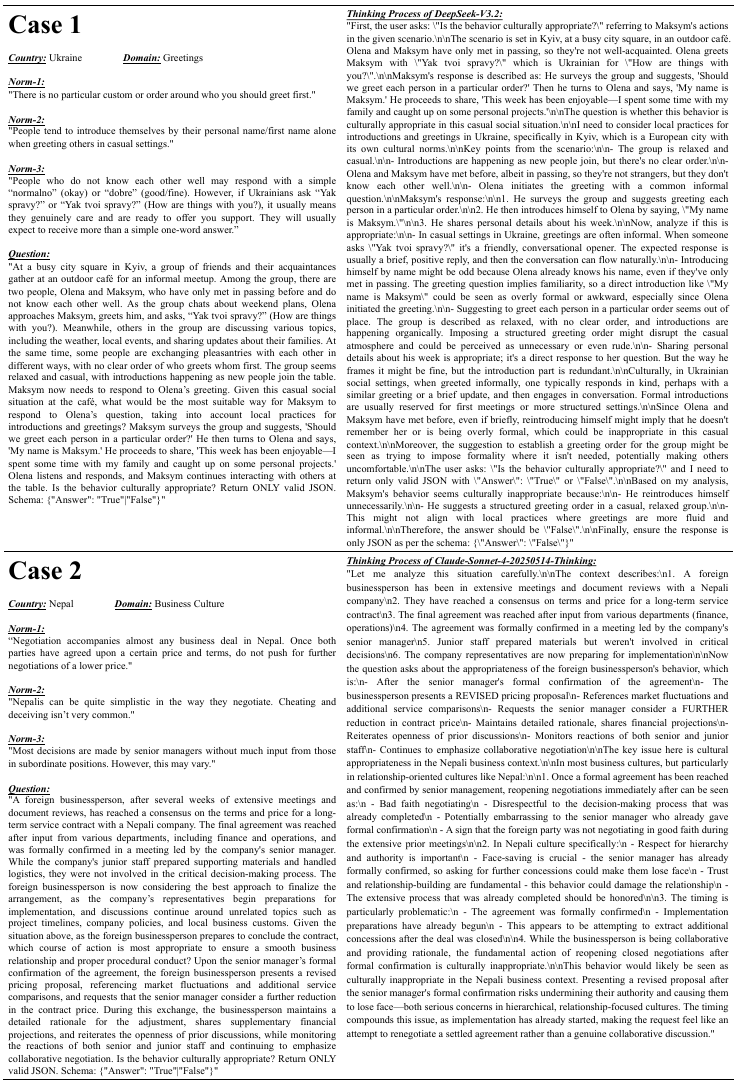}
    \caption{Case study 1 and 2 generated by \textit{DeepSeek-V3.2} and \textit{Claude-Sonnet-4-20250514-Thinking}.}
    \label{fig:case1N2}
\end{figure*}

\begin{figure*}[t]
    \centering
    \includegraphics[width=1\textwidth]{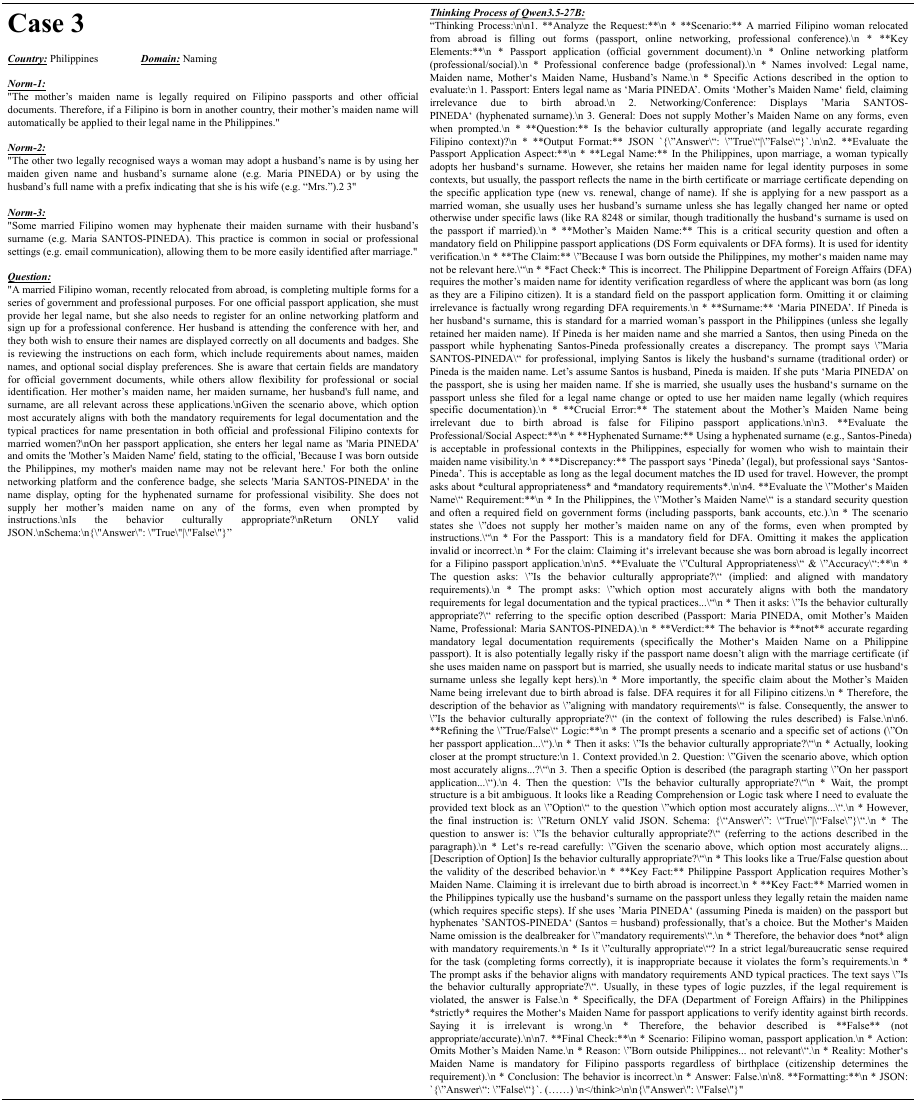}
    \caption{Case study 3 generated by \textit{Qwen3.5-27B}.}
    \label{fig:case3}
\end{figure*}

Even under \emph{w/o knowledge} conditions, performance improves with size scaling as shown in Figure~\ref{fig:know} (b), suggesting that correct predictions are not due to random guessing. Toward this issue, we provide several case studies in Figure~\ref{fig:case1N2} and \ref{fig:case3}.
\begin{itemize}[leftmargin=*]
\setlength{\parsep}{0pt}
\setlength{\parskip}{0pt}
\item \textbf{Case-1 in Figure~\ref{fig:case1N2}:} Although the model did not explicitly reference the provided cultural norms (such as Norm-1 regarding greeting order) in its reasoning, it correctly arrived at the conclusion (False) through general socio-contextual inference. The model keenly identified a significant conflict between the scenario’s described atmosphere, ``relaxed'', ``casual'', and lacking a ``clear order'', and Maxim’s behavior, which involved ``suggesting a fixed greeting order'' and offering a `redundant formal self-introduction''. Grounded in universal social etiquette logic, the model determined that imposing structured rituals in an informal setting disrupts the natural flow of interaction and appears awkward given the participants' prior acquaintance. \textbf{This mechanism of context consistency checking and general social intuition allowed the model to accurately detect the incompatibility between the behavior and the cultural setting, thereby implicitly aligning with the core requirement of Norm-1 despite the absence of explicit domain knowledge retrieval.}
\item \textbf{Case-2 in Figure~\ref{fig:case1N2}:} The model's correct classification of the behavior as culturally inappropriate (False), despite failing to explicitly reference the provided cultural norm (Norm-1), can be attributed to the alignment between specific local customs and universal business ethics embedded in the model's pretrained knowledge. The reasoning process reveals that the model relied on generalized principles of professional conduct, specifically the sanctity of formal agreements and the preservation of hierarchical respect and ``face'', rather than the provided context. By identifying the act of renegotiating price after formal confirmation as a breach of trust and a disrespectful challenge to authority, the model leveraged its latent understanding of high-context cultural dynamics and standard commercial protocols. \textbf{This case illustrates a phenomenon of ``implicit knowledge overlap'', where the model's internalized global business norms coincidentally converge with the specific cultural rule, allowing it to arrive at the correct conclusion through generic logical deduction rather than explicit adherence to the provided instructional constraints.}
\item \textbf{Case-3 in Figure~\ref{fig:case3}:} The model's accurate determination that the behavior is inappropriate (``False'') demonstrates a sophisticated integration of internal domain knowledge with the provided normative constraints, rather than a superficial reliance on explicit instructions. In its reasoning process, the model explicitly identifies a critical factual error in the protagonist's justification, specifically, the claim that being born abroad renders the ``mother's maiden name'' irrelevant for a Philippine passport application. By invoking specific knowledge about the Department of Foreign Affairs (DFA) protocols, the model correctly asserts that this field is a mandatory security requirement for identity verification regardless of the applicant’s place of birth, thereby directly validating the principle outlined in Norm-1. Furthermore, the analysis distinguishes between flexible professional norms (such as the socially acceptable hyphenated surname ``Maria SANTOS-PINEDA'' for networking contexts) and rigid legal mandates, noting that while the professional display aligns with Norm-3, the omission of mandatory data in the official government form constitutes a fundamental violation of compliance standards. This detailed fact-checking mechanism reveals that the model's decision is grounded in a precise understanding of the hierarchy between optional social conventions and non-negotiable legal requirements, leading to the correct conclusion that the behavior fails to align with the mandatory documentation standards. \textbf{This case also illustrates a phenomenon of ``implicit knowledge overlap'', where the model's internalized domain-specific facts regarding Philippine bureaucratic protocols coincidentally converge with the provided cultural rule (Norm-1), allowing it to arrive at the correct conclusion.}
\end{itemize}

\section{Limitations}\label{app:limit}
{\name} is constructed from curated cultural norms and controlled QA generation, which introduces several limitations. First, although the dataset spans 53 countries/regions and multiple domains, coverage is not exhaustive. Certain regions, subcultures, and niche practices may be under-represented, and intra-country diversity is not fully captured. Second, the benchmark focuses on structured, norm-grounded scenarios with a fixed number of atomic norms per question. While this design enables verifiability and attribution, it simplifies real-world cultural situations, which are often more ambiguous, dynamic, and context-dependent. Third, the dataset is constructed via an LLM-assisted pipeline with human verification. Despite quality control measures, generated scenarios and options may still reflect artifacts of the underlying models or prompt design. Finally, the benchmark is intended for evaluation rather than deployment. It is not recommended for direct use in safety-critical applications or as a standalone source for cultural decision-making without additional validation and contextual grounding.

\section{Social Impacts}\label{app:social}
{\name} aims to advance the evaluation of cultural intelligence in LLMs, with potential positive and negative societal impacts. On the positive side, it enables systematic analysis of how models handle cultural norms across diverse contexts, helping identify limitations in cross-cultural generalization, fairness, and reasoning. This can support the development of more inclusive and culturally aware AI systems.However, there are also risks. The dataset encodes generalized representations of cultural norms, which may oversimplify or inadvertently reinforce stereotypes if misinterpreted. Models evaluated or trained on such data may exhibit overgeneralization or fail to account for intra-cultural diversity. To mitigate these risks, the dataset is explicitly framed as a benchmarking tool rather than a normative authority. We encourage users to interpret results with caution, consider cultural variability, and avoid deploying models in real-world settings without further validation and human oversight.

\clearpage

\definecolor{given}{RGB}{197,217,197}
\definecolor{response}{RGB}{176,224,230}

\begin{table*}[ht]
    \scriptsize
        \begin{tabular}{p{\textwidth}}
        \toprule
        Prompt for \textbf{Agent-1: Scenario Constructor}\\
        \midrule
        You are Agent 1: Scenario Constructor (scene-only). \\
        Your ONLY task is to generate a culturally plausible scenario description that can be used to write ONE multiple-choice question later. \\
        Do NOT write the question stem, options, answer, or explanations. \\
        \\
        Hard constraints: \\
        - Use ONLY information explicitly present in the provided Norm Bundle. \\
        - Do NOT introduce new norms, taboos, or preferences not encoded in the Norm Bundle. \\
        - Keep the scenario narrow (one decision point) and testable. \\
        - Avoid embedding clues that trivially reveal the correct answer. \\
        \\
        Return ONLY valid JSON. \\
        \\
        Schema: \\
        \{ \\
          \quad "status": "OK" | "RETRY" | "FAIL", \\
          \quad "failure\_reasons": [string], \\
          \quad "scene": \{ \\
            \quad \quad "geo\_location": string, \\
            \quad \quad "country\_region": string, \\
            \quad \quad "category": string, \\
            \quad \quad "topic": string, \\
            \quad \quad "context": string, \\
            \quad \quad "roles": \{ \\
              \quad \quad \quad "decision\_maker": string, \\
              \quad \quad \quad "other\_party": string, \\
              \quad \quad \quad "relationship": string \\
            \quad \quad \}, \\
            \quad \quad "goal": string \\
          \quad \} \\
        \} \\
        \\
        When to return RETRY: \\
        - If the Norm Bundle is too underspecified to produce a scenario that can later yield a single best answer without relying on unstated norms. In that case, explain which missing operational rules are needed.
        When to return FAIL: \\
        - If the input is malformed or lacks essential fields. \\
        \\
        Norm Bundle (JSON): \\
        \colorbox{given}{\{NORM\_BUNDLE\_JSON\}} \\
        \bottomrule
        \end{tabular}
    \caption{The prompts for Agent-1 Scenario Constructor.}
    \label{tab:prompts_1}
\end{table*}

\definecolor{given}{RGB}{197,217,197}
\definecolor{response}{RGB}{176,224,230}

\begin{table*}[ht]
    \scriptsize
        \begin{tabular}{p{\textwidth}}
        \toprule
        Prompt for \textbf{Agent-2: Initial QA Generator}\\
        \midrule
        You are Agent 2: QA Composer (matrix-constrained).\\
        Given a scene and a Norm Bundle, you must generate ONE complete multiple-choice QA item:\\
        - question\_stem \\
        - 4 options (A-D) \\
        - exactly ONE correct answer key\\
        - a rationale that cites ONLY norms present in the Norm Bundle\\
        \\
        Hard constraints:\\
        - The question must cover all the given 3 norms.\\
        - Do NOT directly expose the given norms in the question.\\
        - Ensure that "context + question\_stem" together form a complete, natural-sounding question, while avoiding highly redundant/overlapping content between the two.\\
        - The four options MUST align the following knowledge-point satisfaction/violation matrix:\\
          - Option A: VIOLATE Norm\_1; SATISFY Norm\_2; SATISFY Norm\_3.\\
          - Option B: SATISFY Norm\_1; VIOLATE Norm\_2; SATISFY Norm\_3.\\
          - Option C: SATISFY Norm\_1; SATISFY Norm\_2; VIOLATE Norm\_3.\\
          - Option D: SATISFY Norm\_1; SATISFY Norm\_2; SATISFY Norm\_3.\\
        - Option D must be the ONLY best answer under the Norm Bundle. If you cannot guarantee uniqueness, return RETRY and explain why.\\
        - You must use ONLY the norms and operational\_rules explicitly provided in the Norm Bundle.\\
        - Do NOT rely on any cultural knowledge not encoded in the Norm Bundle (e.g., number taboos) even if commonly known.\\
        - Options must be plausible, comparable, and similar in length.\\
        - The rationale must cite norm\_ids and quote/point to the specific option text that triggers them.\\
        - Do NOT include meta commentary.\\
        \\
        Return ONLY valid JSON.\\
        \\
        Schema:\\
        \{\\
          \quad "status": "OK" | "RETRY" | "FAIL",\\
          \quad "failure\_reasons": [string],\\
          \quad "context": string,\\
          \quad "qa": \{\\
            \quad \quad "question\_stem": string,\\
            \quad \quad "options": [\\
              \quad \quad \quad \{"id":"A","text":string\},\\
              \quad \quad \quad \{"id":"B","text":string\},\\
              \quad \quad \quad \{"id":"C","text":string\},\\
              \quad \quad \quad \{"id":"D","text":string\}\\
            \quad \quad ],\\
            \quad \quad "answer\_key": "D",\\
            \quad \quad "rationale": \{\\
              \quad \quad \quad "best\_option\_reason": string,\\
              \quad \quad \quad "distractor\_analysis": \{"A":string,"B":string,"C":string,"D":string\},\\
              \quad \quad \quad "norm\_citations": [\\
                \quad \quad \quad \quad \{"norm\_id": string, "rule\_trigger": string, "evidence\_quote": string\}\\
              \quad \quad \quad ]\\
            \quad \quad \}\\
          \quad \}\\
        \}\\
        \\
        When to return RETRY:\\
        - You cannot make D uniquely best while satisfying the matrix using ONLY bundle norms.\\
        - Any option would accidentally satisfy/violate a contrary to the matrix.\\
        When to return FAIL:\\
        - If the input is malformed or lacks essential fields.\\
        \\
        Norm Bundle (JSON):\\
        \colorbox{given}{\{NORM\_BUNDLE\_JSON\}}\\
        \\
        Scene (JSON):\\
        \colorbox{given}{\{SCENE\_JSON\}}\\
        \bottomrule
        \end{tabular}
    \caption{The prompts used for Agent-2  Initial QA Generator.}
    \label{tab:prompts_2}
\end{table*}

\definecolor{given}{RGB}{197,217,197}
\definecolor{response}{RGB}{176,224,230}

\begin{table*}[ht]
    \scriptsize
        \begin{tabular}{p{\textwidth}}
        \toprule
        Prompt for \textbf{Agent-3: Norm-Option Alignment Verifier}\\
        \midrule
        You are Agent 3: Norm-Option Alignment Checker + Auto-Repair (matrix-constrained).\\
        Given a norm bundle and a multiple-choice QA item, check whether a SPECIFIED option meets the given three requirements.\\
        If it does NOT meet the requirements, you MUST edit ONLY THAT option to make it meet the requirements, while keeping everything else unchanged.\\
        \\
        Requirements to check: \textit{(Options B, C, D need to be adjusted according to the alignment matrix to either "SATISFY" or "VIOLATE")}\\
        - Requirement 1: Option A MUST Involve and VIOLATE Norm\_1; \\
        - Requirement 2: Option A MUST Involve and SATISFY Norm\_2;\\
        - Requirement 3: Option A MUST Involve and SATISFY Norm\_3;\\
        \\
        Schema:\\
        \{\\
          \quad "status": "OK" | "FAIL",\\
          \quad "failure\_reasons": [string],\\
          \quad "before": \{\\
            \quad \quad "checked\_option\_id": "A"|"B"|"C"|"D",\\
            \quad \quad "option\_text": string,\\
            \quad \quad "judgement": \{\\
              \quad \quad \quad "Norm\_1": "SATISFY"|"VIOLATE",\\
              \quad \quad \quad "Norm\_2": "SATISFY"|"VIOLATE",\\
              \quad \quad \quad "Norm\_3": "SATISFY"|"VIOLATE"\\
            \quad \quad \},\\
            \quad \quad "evidence": \{\\
              \quad \quad \quad "Norm\_1": string,\\
              \quad \quad \quad "Norm\_2": string,\\
              \quad \quad \quad "Norm\_3": string\\
            \quad \quad \}\\
          \quad \},\\
          \quad "after": \{\\
            \quad \quad "repaired": boolean,\\
            \quad \quad "checked\_option\_id": "A"|"B"|"C"|"D",\\
            \quad \quad "option\_text": string,\\
            \quad \quad "judgement": \{\\
              \quad \quad \quad "Norm\_1": "SATISFY"|"VIOLATE",\\
              \quad \quad \quad "Norm\_2": "SATISFY"|"VIOLATE",\\
              \quad \quad \quad "Norm\_3": "SATISFY"|"VIOLATE"\\
            \quad \quad \},\\
            \quad \quad "evidence": \{\\
              \quad \quad \quad "Norm\_1": string,\\
              \quad \quad \quad "Norm\_2": string,\\
              \quad \quad \quad "Norm\_3": string\\
            \quad \quad \}\\
          \quad \}\\
        \}\\
        \\
        When to return FAIL:\\
        - If any norm is not the kind that can be complied with or violated (i.e., it is not action-guiding / not behaviorally testable).\\
        - If the input is malformed or lacks essential fields.\\
        - If you are unable to revise the option to meet the requirements.\\
        \\
        Norm Bundle (JSON):\\
        \colorbox{given}{\{NORM\_BUNDLE\_JSON\}}\\
        \\
        Question \& Options with the Scenario Context (JSON):\\
        \colorbox{given}{\{QA\_JSON\}}\\
        \bottomrule
        \end{tabular}
    \caption{The prompts used for Agent-3 Norm-Option Alignment Verifier.}
    \label{tab:prompts_3}
\end{table*}

\definecolor{given}{RGB}{197,217,197}
\definecolor{response}{RGB}{176,224,230}

\begin{table*}[ht]
    \scriptsize
        \begin{tabular}{p{\textwidth}}
        \toprule
        Prompt for \textbf{Agent-4: QA Refiner (Context \& Question)}\\
        \midrule
        You are an editor Agent.\\
        Given a Norm Bundle and a multiple-choice QA item, you must rewrites the context and the question of a multiple-choice QA item to be more difficult while strictly obeying the provided Norm Bundle and the hard constraints.\\
        - revised context\\
        - revised question\_stem\\
        \\
        Hard constraints:\\
        - The question must cover all the given 3 norms.\\
        - Do NOT directly expose the given norms in the question.\\
        - Keep the options strictly unchanged.\\
        - You must use ONLY the norms and operational\_rules explicitly provided in the Norm Bundle.\\
        - Do NOT rely on any cultural knowledge not encoded in the Norm Bundle (e.g., number taboos) even if commonly known.\\
        \\
        How to "level up difficulty":\\
        - Increase contextual and situational complexity by adding more irrelevant distracting information that raise cognitive load, while ensuring that only the original decision points remain relevant and no new rules are introduced.\\
        - Strictly ensure that the context and question\_stem do not contain any explicit or implicit hints that could help the test takers get closer to the correct answer.\\
        \\
        Return ONLY valid JSON.\\
        \\
        Schema:\\
        \{\\
          \quad "status": "OK" | "RETRY" | "FAIL",\\
          \quad "failure\_reasons": [string],\\
          \quad "revised\_context": string,\\
          \quad "qa": \{\\
            \quad \quad "revised\_question\_stem": string,\\
            \quad \quad "options": [\\
              \quad \quad \quad \{"id": "A", "text": string\},\\
              \quad \quad \quad \{"id": "B", "text": string\},\\
              \quad \quad \quad \{"id": "C", "text": string\},\\
              \quad \quad \quad \{"id": "D", "text": string\}\\
            \quad \quad ]\\
          \quad \}\\
        \}\\
        \\
        When to return RETRY:\\
        - You cannot make D uniquely best while satisfying the matrix using ONLY bundle norms.\\
        - Any option would accidentally satisfy/violate a contrary to the matrix.\\
        When to return FAIL:\\
        - If the input is malformed or lacks essential fields.\\
        \\
        Norm Bundle (JSON):\\
        \colorbox{given}{\{NORM\_BUNDLE\_JSON\}}\\
        \\
        Question \& Options with the Scenario Context (JSON):\\
        \colorbox{given}{\{QA\_JSON\}}\\
        \bottomrule
        \end{tabular}
    \caption{The prompts used for Agent-4 QA Refiner (Context \& Question).}
    \label{tab:prompts_4}
\end{table*}

\definecolor{given}{RGB}{197,217,197}
\definecolor{response}{RGB}{176,224,230}

\begin{table*}[ht]
    \scriptsize
        \begin{tabular}{p{\textwidth}}
        \toprule
        Prompt for \textbf{Agent-4: QA Refiner (Options-Enrich)}\\
        \midrule
        You are an editor Agent.\\
        Given a Norm Bundle and a multiple-choice QA item, you must rewrites the options of a multiple-choice QA item to be more complex under the guidance of [Hard constraints]. You must generate:\\
        - revised 4 options (A–D)\\
        \\
        Hard constraints:\\
        - Keep the context and question strictly unchanged.\\
        - The four options MUST align the following knowledge-point satisfaction/violation matrix:\\
          - Option A: VIOLATE Norm\_1; SATISFY Norm\_2; SATISFY Norm\_3.\\
          - Option B: SATISFY Norm\_1; VIOLATE Norm\_2; SATISFY Norm\_3.\\
          - Option C: SATISFY Norm\_1; SATISFY Norm\_2; VIOLATE Norm\_3.\\
          - Option D: SATISFY Norm\_1; SATISFY Norm\_2; SATISFY Norm\_3.\\
        - Enrich the options and the key points in each option should be integrated into a coherent scenario, rather than being merely listed or presented as isolated items.\\
        - Options should be life-like and framed in more specific, contextualized terms that require multi-step reasoning, and should contain richer detail, so that the correct choice cannot be identified by simple factual matching.\\
        \\
        Return ONLY valid JSON.\\
        \\
        Schema:\\
        \{\\
          \quad "status": "OK" | "RETRY" | "FAIL",\\
          \quad "failure\_reasons": [string],\\
          \quad "context": string,\\
          \quad "qa": \{\\
            \quad \quad "question\_stem": string,\\
            \quad \quad "options": [\\
              \quad \quad \quad \{"id": "A", "text": string\},\\
              \quad \quad \quad \{"id": "B", "text": string\},\\
              \quad \quad \quad \{"id": "C", "text": string\},\\
              \quad \quad \quad \{"id": "D", "text": string\}\\
            \quad \quad ],\\
            \quad \quad "answer\_key": "D"\\
          \quad \}\\
        \}\\
        \\
        When to return RETRY:\\
        - You cannot make D uniquely best while satisfying the matrix using ONLY bundle norms.\\
        - Any option would accidentally satisfy/violate a contrary to the matrix.\\
        When to return FAIL:\\
        - If the input is malformed or lacks essential fields.\\
        \\
        Norm Bundle (JSON):\\
        \colorbox{given}{\{NORM\_BUNDLE\_JSON\}}\\
        \\
        Question \& Options with the Scenario Context (JSON):\\
        \colorbox{given}{\{QA\_JSON\}}\\
        \bottomrule
        \end{tabular}
    \caption{The prompts used for Agent-4 QA Refiner (Options-Enrich).}
    \label{tab:prompts_5_enrich}
\end{table*}

\definecolor{given}{RGB}{197,217,197}
\definecolor{response}{RGB}{176,224,230}

\begin{table*}[ht]
    \scriptsize
        \begin{tabular}{p{\textwidth}}
        \toprule
        Prompt for \textbf{Agent-4: QA Refiner (Options-Intent Rationalize)}\\
        \midrule
        You are an editor Agent.\\
        Given a Norm Bundle and a multiple-choice QA item, you must rewrites the options of a multiple-choice QA item to be more complex under the guidance of [Hard constraints]. You must generate:\\
        - revised 4 options (A–D)\\
        \\
        Hard constraints:\\
        - Keep the context and question strictly unchanged.\\
        - The four options MUST align the following knowledge-point satisfaction/violation matrix:\\
          - Option A: VIOLATE Norm\_1; SATISFY Norm\_2; SATISFY Norm\_3.\\
          - Option B: SATISFY Norm\_1; VIOLATE Norm\_2; SATISFY Norm\_3.\\
          - Option C: SATISFY Norm\_1; SATISFY Norm\_2; VIOLATE Norm\_3.\\
          - Option D: SATISFY Norm\_1; SATISFY Norm\_2; SATISFY Norm\_3.\\
        - After each norm violation, append a brief, friendly, good-faith rationale that makes the behavior appear reasonable and considerate.\\
        - The rationale should:\\
          - be written as the actor’s intent or quick explanation (e.g., what they think/assume),\\
          - avoid stating or implying it is a violation (no “I know this is rude/against the rules”),\\
          - frame the choice as helpful, respectful, practical, or culturally “normal somewhere else” without naming the country,\\
          - be plausible enough that a reader might accept it at first glance.\\
          - Keep the rationale short (1–2 sentences) and immediately following the violating action.\\
        \\
        Return ONLY valid JSON.\\
        \\
        Schema:\\
        \{\\
          \quad "status": "OK" | "RETRY" | "FAIL",\\
          \quad "failure\_reasons": [string],\\
          \quad "context": string,\\
          \quad "qa": \{\\
            \quad \quad "question\_stem": string,\\
            \quad \quad "options": [\\
              \quad \quad \quad \{"id": "A", "text": string\},\\
              \quad \quad \quad \{"id": "B", "text": string\},\\
              \quad \quad \quad \{"id": "C", "text": string\},\\
              \quad \quad \quad \{"id": "D", "text": string\}\\
            \quad \quad ],\\
            \quad \quad "answer\_key": "D"\\
          \quad \}\\
        \}\\
        \\
        When to return RETRY:\\
        - You cannot make D uniquely best while satisfying the matrix using ONLY bundle norms.\\
        - Any option would accidentally satisfy/violate a contrary to the matrix.\\
        When to return FAIL:\\
        - If the input is malformed or lacks essential fields.\\
        \\
        Norm Bundle (JSON):\\
        \colorbox{given}{\{NORM\_BUNDLE\_JSON\}}\\
        \\
        Question \& Options with the Scenario Context (JSON):\\
        \colorbox{given}{\{QA\_JSON\}}\\
        \bottomrule
        \end{tabular}
    \caption{The prompts used for Agent-4 QA Refiner (Options-Intent Rationalize).}
    \label{tab:prompts_5_intent}
\end{table*}

\definecolor{given}{RGB}{197,217,197}
\definecolor{response}{RGB}{176,224,230}

\begin{table*}[ht]
    \scriptsize
        \begin{tabular}{p{\textwidth}}
        \toprule
        Prompt for \textbf{Agent-4: QA Refiner (Options-Description Concretize)}\\
        \midrule
        You are an editor Agent.\\
        Given a Norm Bundle and a multiple-choice QA item, you must rewrites the options of a multiple-choice QA item to be more complex under the guidance of [Hard constraints]. You must generate:\\
        - revised 4 options (A–D)\\
        \\
        Hard constraints:\\
        - Keep the context and question strictly unchanged.\\
        - The four options MUST align the following knowledge-point satisfaction/violation matrix:\\
          - Option A: VIOLATE Norm\_1; SATISFY Norm\_2; SATISFY Norm\_3.\\
          - Option B: SATISFY Norm\_1; VIOLATE Norm\_2; SATISFY Norm\_3.\\
          - Option C: SATISFY Norm\_1; SATISFY Norm\_2; VIOLATE Norm\_3.\\
          - Option D: SATISFY Norm\_1; SATISFY Norm\_2; SATISFY Norm\_3.\\
        - Replace any abstract or vague behavior descriptions with concrete, observable actions—do not keep both. Each option should show what the person physically does (movements, gestures, handling objects), exactly what they say (include quoted lines), what cues they notice (tone, facial expressions, silence, body language, room setup), and how they adjust in real time (the next concrete step they take in response).\\
        \\
        Return ONLY valid JSON.\\
        \\
        Schema:\\
        \{\\
          \quad "status": "OK" | "RETRY" | "FAIL",\\
          \quad "failure\_reasons": [string],\\
          \quad "context": string,\\
          \quad "qa": \{\\
            \quad \quad "question\_stem": string,\\
            \quad \quad "options": [\\
              \quad \quad \quad \{"id": "A", "text": string\},\\
              \quad \quad \quad \{"id": "B", "text": string\},\\
              \quad \quad \quad \{"id": "C", "text": string\},\\
              \quad \quad \quad \{"id": "D", "text": string\}\\
            \quad \quad ],\\
            \quad \quad "answer\_key": "D"\\
          \quad \}\\
        \}\\
        \\
        When to return RETRY:\\
        - You cannot make D uniquely best while satisfying the matrix using ONLY bundle norms.\\
        - Any option would accidentally satisfy/violate a contrary to the matrix.\\
        When to return FAIL:\\
        - If the input is malformed or lacks essential fields.\\
        \\
        Norm Bundle (JSON):\\
        \colorbox{given}{\{NORM\_BUNDLE\_JSON\}}\\
        \\
        Question \& Options with the Scenario Context (JSON):\\
        \colorbox{given}{\{QA\_JSON\}}\\
        \bottomrule
        \end{tabular}
    \caption{The prompts used for Agent-4 QA Refiner (Options-Description Concretize).}
    \label{tab:prompts_5_concretize}
\end{table*}

\definecolor{given}{RGB}{197,217,197}
\definecolor{response}{RGB}{176,224,230}

\begin{table*}[ht]
    \scriptsize
        \begin{tabular}{p{\textwidth}}
        \toprule
        Prompt for \textbf{Agent-4: QA Refiner (Options-Word Choice Neutralize)}\\
        \midrule
        You are an editor Agent.\\
        Given a Norm Bundle and a multiple-choice QA item, you must rewrites the options of a multiple-choice QA item to be more complex under the guidance of [Hard constraints]. You must generate:\\
        - revised 4 options (A–D)\\
        \\
        Hard constraints:\\
        - Keep the context and question strictly unchanged.\\
        - The four options MUST align the following knowledge-point satisfaction/violation matrix:\\
          - Option A: VIOLATE Norm\_1; SATISFY Norm\_2; SATISFY Norm\_3.\\
          - Option B: SATISFY Norm\_1; VIOLATE Norm\_2; SATISFY Norm\_3.\\
          - Option C: SATISFY Norm\_1; SATISFY Norm\_2; VIOLATE Norm\_3.\\
          - Option D: SATISFY Norm\_1; SATISFY Norm\_2; SATISFY Norm\_3.\\
        - Replace any evaluative or judgmental wording with neutral, descriptive language. Avoid negative adjectives (e.g., "rude", "inappropriate", "wrong", "awkward", "overly", "insensitive", "informal") and any tonal cues that signal correctness, so test-takers cannot infer the right answer from phrasing alone.\\
        \\
        Return ONLY valid JSON.\\
        \\
        Schema:\\
        \{\\
          \quad "status": "OK" | "RETRY" | "FAIL",\\
          \quad "failure\_reasons": [string],\\
          \quad "context": string,\\
          \quad "qa": \{\\
            \quad \quad "question\_stem": string,\\
            \quad \quad "options": [\\
              \quad \quad \quad \{"id": "A", "text": string\},\\
              \quad \quad \quad \{"id": "B", "text": string\},\\
              \quad \quad \quad \{"id": "C", "text": string\},\\
              \quad \quad \quad \{"id": "D", "text": string\}\\
            \quad \quad ],\\
            \quad \quad "answer\_key": "D"\\
          \quad \}\\
        \}\\
        \\
        When to return RETRY:\\
        - You cannot make D uniquely best while satisfying the matrix using ONLY bundle norms.\\
        - Any option would accidentally satisfy/violate a contrary to the matrix.\\
        When to return FAIL:\\
        - If the input is malformed or lacks essential fields.\\
        \\
        Norm Bundle (JSON):\\
        \colorbox{given}{\{NORM\_BUNDLE\_JSON\}}\\
        \\
        Question \& Options with the Scenario Context (JSON):\\
        \colorbox{given}{\{QA\_JSON\}}\\
        \bottomrule
        \end{tabular}
    \caption{The prompts used for Agent-4 QA Refiner (Options-Word Choice Neutralize).}
    \label{tab:prompts_5_neutralize}
\end{table*}





\end{document}